\documentclass[sigconf]{acmart}

\setcopyright{acmcopyright}
\copyrightyear{2018}
\acmYear{2018}
\acmDOI{10.1145/1122445.1122456}

\acmConference[ICAIL '21]{ICAIL '21: International Conference on Artificial Intelligence and Law}{June, 21--25, 2021}{São Paulo, Brasil}
\acmBooktitle{Woodstock '18: ACM Symposium on Neural Gaze Detection,
  June 03--05, 2018, Woodstock, NY}
\acmPrice{15.00}
\acmISBN{978-1-4503-XXXX-X/18/06}

\usepackage[pscoord]{eso-pic}%
\definecolor{Preprint}{rgb}{.63,.79,.95}
\newcommand{\preprint}[4]{
\AddToShipoutPictureFG*{\put(\LenToUnit{0.5\paperwidth},\LenToUnit{0.95\paperheight}){\makebox[0pt][c]{
\renewcommand{\arraystretch}{1.5} \setlength{\tabcolsep}{18pt} 
\begin{tabular}{|p{0.75\paperwidth}|} \hline
\rowcolor{Preprint} \textbf{Preprint from \texttt{\href{https://ostendorff.org/pub/}{https://ostendorff.org/pub/}}} \\ 
\rowcolor{Gray} \hline \footnotesize #1. ``#2'' in \textit{#3}, #4. \\ \hline
\end{tabular}}}}}

\renewcommand\footnotetextcopyrightpermission[1]{} %

\usepackage{graphicx}
\usepackage{times}
\usepackage{url}
\usepackage{latexsym}

\usepackage{graphicx}
\usepackage{tabularx,ragged2e}
\usepackage{tabulary}
\usepackage{adjustbox}
\usepackage{amsmath}
\usepackage{multirow}
\usepackage{url}
\usepackage{hhline}
\usepackage{multicol}
\usepackage{booktabs}
\usepackage{caption}
\usepackage{subcaption}
\usepackage[multiple]{footmisc}
\usepackage{xspace}

\usepackage{makecell}

\newcommand{\R}{\mathbb{R}}

\usepackage{color, colortbl}
\definecolor{Gray}{gray}{0.925} %

\newcommand{\extended}[2]{#2}

\newcommand{\methodsCount}{27\xspace} %
\newcommand{\excludedMethodsCount}{14\xspace}  %
\newcommand{\allMethodsCount}{41\xspace} %

\newcommand{\silverCountText}{2,964\xspace} 
\newcommand{\wsCount}{1363\xspace} 
\newcommand{\wsCountText}{1,363\xspace} 

\newcommand{\ocbCount}{1601\xspace} 
\newcommand{\ocbCountText}{1,601\xspace}

\newcommand{\ocb}{Open Case Book\xspace} 
\newcommand{\ws}{Wikisource\xspace} 
\newcommand{\fastText}{fastText\xspace} 
\newcommand{\fastTextLegal}{fastText\-\textsubscript{Legal}\xspace} 
\newcommand{\legalJHU}{Legal-JHU-BERT\xspace} 
\newcommand{\legalAUEB}{Legal-AUEB-BERT\xspace} 
\newcommand{\gloVe}{GloVe\xspace} 
\newcommand{\gloVeLegal}{GloVe\textsubscript{Legal}\xspace} 
\newcommand{\poincare}{Poincar\'{e}\xspace} 
\newcommand{\poincareSumFastTextLegal}{Poincar\'{e} + fastText\-\textsubscript{Legal}\xspace} 
\newcommand{\poincareCatFastTextLegal}{Poincar\'{e} $\mathbin\Vert$ fastText\-\textsubscript{Legal}\xspace} 
\newcommand{\longformerCatFastTextLegal}{Longformer-large $\mathbin\Vert$ fastText\-\textsubscript{Legal}\xspace} 
 
\newcommand{\poincareSumLongformer}{Poincar\'{e} + Longformer-large\xspace}

\begin{document}

\title{Evaluating Document Representations for Content-based Legal Literature Recommendations}

\author{Malte Ostendorff${}^{1,2}$, Elliott Ash${}^{3}$, Terry Ruas${}^{4}$, Bela Gipp${}^{4}$, Julian Moreno-Schneider${}^{2}$, Georg Rehm${}^{2}$}

\affiliation{
	\institution{
    \textsuperscript{1}Open Legal Data
    \country{Germany} 
    (mo@openlegaldata.io)
    }
}

\affiliation{
	\institution{
    \textsuperscript{2}German Research Center for Artificial Intelligence
    \country{Germany}
    (firstname.lastname@dfki.de)
    }
}

\affiliation{
	\institution{
    \textsuperscript{3}ETH Zurich
    \country{Switzerland}
    (ashe@ethz.ch)
    }
}

\affiliation{
	\institution{
    \textsuperscript{4}University of Wuppertal
    \country{Germany}
    (lastname@uni-wuppertal.de)
    }
}
\renewcommand{\shortauthors}{Ostendorff et al.}

\begin{abstract}

Recommender systems assist legal professionals in finding relevant literature for supporting their case.
Despite its importance for the profession, legal applications do not reflect the latest advances in recommender systems and representation learning research. 
Simultaneously, legal recommender systems are typically evaluated in small-scale user study without any public available benchmark datasets.
Thus, these studies have limited reproducibility. 
To address the gap between research and practice, we explore a set of state-of-the-art document representation methods for the task of retrieving semantically related US case law.
We evaluate text-based (e.g., fastText, Transformers), citation-based (e.g., DeepWalk, \poincare), and hybrid methods. 
We compare in total \methodsCount methods using two silver standards with annotations for \silverCountText documents.
The silver standards are newly created from \ocb and \ws and can be reused under an open license facilitating reproducibility.
Our experiments show that document representations from averaged \fastText word vectors (trained on legal corpora) yield the best results, closely followed by \poincare citation embeddings. 
Combining \fastText and \poincare in a hybrid manner further improves the overall result. 
Besides the overall performance, we analyze the methods depending on document length, citation count, and the coverage of their recommendations.
We make our source code, models, and datasets publicly available.

\end{abstract}



\maketitle

\preprint{Malte Ostendorff, Elliott Ash, Terry Ruas, Bela Gipp, Julian Moreno-Schneider, Georg Rehm}{Evaluating Document Representations for Content-based Legal Literature Recommendations}{Proceedings of the 18th International Conference on Artificial Intelligence and Law (ICAIL 2021)}{2021}

\section{Introduction}

Legal professionals, e.g., lawyers and judges, frequently invest considerable time to find relevant literature~\cite{Lastres2013}.
More so than most other domains, in law there are high stakes for finding the most relevant information (documents) as that can drastically affect the outcome of a dispute. 
A case can be won or lost depending on whether or not a supporting decision can be found.
Recommender systems assist in the search for relevant information.
However, research and development of recommender systems for legal corpora poses several challenges. 
Recommender system research is known to be domain-specific, i.e., minor changes may lead to unpredictable variations in the recommendation effectiveness~\cite{Beel2016}.
Likewise, legal English is a peculiarly obscure and convoluted variety of English with a widespread use of common words with uncommon meanings~\cite{Mellinkoff1963}.
Recent language models like BERT~\cite{Devlin2019} may not be equipped to handle legal English since they are pretrained on generic corpora like Wikipedia or cannot process lengthy legal documents due to their limited input length. 
This raises the question of whether the recent advances in recommender system research and underlying techniques are also applicable to law.

In this paper, we empirically evaluate \methodsCount document representation methods and analyze the results with respect to the aforementioned possible issues. 
In particular, we evaluate for each method the quality of the document representations in a literature recommender use case.
The methods are distinguished in three categories: 
(1) word vector-based, 
(2) Transformer-based, 
and (3) citation-based methods.
Moreover, we test additional hybrid variations of the aforementioned methods.
Our primary evaluation metric comes from two silver standards on US case law that we extract from \ocb and \ws.
The relevance annotations from the silver standards are provided for \silverCountText documents.

In summary, our contributions are: 
(1) We propose and make available two silver standards as benchmarks for legal recommender system research that currently do not exist.
(2) We evaluate \methodsCount methods of which the majority have never been investigated in the legal context with a quantitative study and validate our results qualitatively. 
(3) We show that the hybrid combination of text-based and citation-based methods can further improve the experimental results.

\section{Related Work}
\label{sec:related}

Recommender systems are a well-established research field~\cite{bai2019scientific} but relatively few publications focus on law as the application domain.
Winkels et al.~\cite{Winkels2014} are among the first to present a content-based approach to recommend legislation and case law.
Their system uses the citation graph of Dutch Immigration Law and is evaluated with a user study conducted with three participants.
Boer and Winkels~\cite{Boer2016} propose and evaluate Latent Dirichlet Allocation (LDA) \cite{blei2003latent} as a solution to the cold start problem in collaborative filtering recommender system.
In an experiment with 28 users, they find the user-based approach outperforms LDA.
Wiggers and Verberne~\cite{Wiggers2019} study citations for legal information retrieval and suggest citations should be combined with other techniques to improve the performance. 

Kumar et al.~\cite{Kumar2011} compare four different methods to measure the similarity of Indian Supreme Court decision: 
TF-IDF~\cite{Salton1975} on all document terms, TF-IDF on only specific terms from a legal dictionary, Co-Citation, and Bibliographic Coupling. 
They evaluate the similarity measure on 50 document pairs with five legal domain experts. 
In their experiment, Bibliographic Coupling and TF-IDF on legal terms yield the best results. Mandal et al.~\cite{Mandal2017} extend this work by evaluating LDA and document embeddings (Paragraph Vectors \cite{Le2014}) on the same dataset, whereby Paragraph Vectors was found to correlate the most with the expert annotations.
Indian Supreme Court decisions are also used as evaluation by Wagh and Anand~\cite{Wagh2020}, where they use document similarity based on concepts instead of full-text. 
They extract concepts (groups of words) from the decisions and compute the similarity between documents based on these concepts. Their vector representation, an average of word embeddings and TF-IDF, shows IDF for weighting word2vec embeddings improve results. 
Also, Bhattacharya et al.~\cite{Bhattacharya2020} compare citation similarity methods, i.e., Bibliographic Coupling, Co-citation, Dispersion~\cite{Minocha2015} and Node2Vec~\cite{Grover2016}), and text similarity methods like Paragraph Vectors. 
They evaluate the algorithms and their combinations using a gold standard of 47 document pairs. 
A combination of Bibliographic Coupling and Paragraph Vectors achieves the best results.

With Eunomos, Boella et al.~\cite{Boella2016} present a legal document and knowledge management system that allows searching legal documents. 
The document similarity problem is handled using TF-IDF and cosine similarity.
Other experiments using embeddings for document similarity include \citet{Landthaler2016}, \citet{Nanda2019}, and \citet{Ash2018}.

Even though different methods have been evaluated in the legal domain, most results are not coherent and rely on small-scale user studies. 
This finding emphasizes the need for a standard benchmark to enable reproducibility and comparability~\cite{Beel2016}.
Moreover, the recent Transformer models~\cite{Vaswani2017} or novel citation embeddings have not been evaluated in legal recommendation research.

\section{Methodology}
\label{sec:methodology}

In this section, we describe our quantitative evaluation of \methodsCount methods for legal document recommendations. 
We define the recommendation scenario as follows:
The user, a legal professional, needs to research a particular decision, e.g., to prepare a litigation strategy.
Based on the decision at hand, the system recommends other decisions to its users such that the research task is easy to accomplish.
The recommendation is relevant when it covers the same topic or provides essential background information, e.g., it overruled the seed decision~\cite{VanOpijnen2017}.

\subsection{Case Corpus and Silver Standard}
\label{ssec:dataset}

Most of the previous works (Section~\ref{sec:related}) evaluate recommendation relevance by asking domain experts to provide subjective annotations ~\cite{Boer2016,Kumar2011,Mandal2017,Winkels2014}.
Especially in the legal domain, these expert annotations are costly to collect and, therefore, their quantity is limited. 
For the same reason, expert annotations are rarely published.
Consequently, the research is difficult to reproduce~\cite{Beel2016}.
In the case of the US court decisions, such expert annotations between documents are also not publicly available.
We construct two ground truth datasets from publicly available resources allowing the evaluation of more recommendations to mitigate the mentioned problems of cost, quantity, and reproducibility.

\subsubsection{\ocb}

With \ocb, the Harvard Law School Library offers a platform for making and sharing open-licensed casebooks~\footnote{\label{fn:ocb}\url{https://opencasebook.org}}. %
The corpus consists of 222 casebooks containing 3,023 cases from 87 authors.
Each casebook contains a manually curated set of topically related court decisions, which we use as relevance annotations.
The casebooks cover a range from broad topics (e.g., \textit{Constitutional law}) to specific ones (e.g., \textit{Intermediary Liability and Platforms' Regulation}).
The decisions are mapped to full-texts and citations retrieved from the Caselaw Access Project (CAP)\footnote{\label{fn:cap}\url{https://case.law}}. %
After duplicate removal and the mapping procedure, relevance annotations for \ocbCountText decisions remain.

\subsubsection{\ws}
We use a collection of 2,939 US Supreme Court decisions from \ws as ground truth~\cite{WikiSource2020}. 
The collection is categorized in 67 topics like \emph{antitrust}, \emph{civil rights}, and \emph{amendments}.
We map the decisions listed in \ws to the corpus from CourtListener\footnote{\label{fn:cl}\url{https://courtlistner.com}}. %
The discrepancy between the two corpora decreases the number of relevance annotations to \wsCountText court decisions.

\begin{table}[!ht]
\small
\caption{\label{tab:dataset_stats}Distribution of relevant annotations for \ocb and \ws.}

\setlength{\tabcolsep}{3pt} %
\renewcommand{\arraystretch}{1.} %
 
\begin{tabular}{lrrrrrrrrr}
\toprule
{} & \multicolumn{7}{c}{\textbf{Relevant annotations per document}} \\
 \cmidrule(lr){2-8}
{} 
&    Mean
&    Std. 
&  Min. 
&   25\% 
&    50\% 
&    75\% 
&     Max. \\
\midrule
\ocb 
&   86.42 &  65.18 &  2.0 &  48.0 &   83.0 &  111.0 &  1590.0 \\
\ws   
&  130.01 &  82.46 &  1.0 &  88.0 &  113.0 &  194.0 &   616.0 \\
\bottomrule
\end{tabular}

\end{table}

We derive a binary relevance classification from \ocb and \ws.
When decisions A and B are in the same casebook or category, A is relevant for B and vice versa.
Table~\ref{tab:dataset_stats} presents the distribution of relevance annotations.
This relevance classification is limited since a recommendation might still be relevant despite not being assigned to the same topic as the seed decision.
Thus, we consider the \ocb and \ws annotations as a silver standard rather than a gold one.

\subsection{Evaluated Methods}
We evaluate \methodsCount methods, each representing legal document $d$ as a numerical vector $\vec{d} \in \R^s$, with $s$ denoting the vector size.
To retrieve the recommendations, we first obtain the vector representations (or document embeddings).
Next, we compute the cosine similarities of the vectors. 
Finally, we select the top $k=5$ documents with the highest similarity through nearest neighbor search\footnote{We set $k=5$ due to the UI \cite{Ostendorff2020b} into which the recommendations will be integrated.}.
Mean Average Precision (MAP) is the primary and Mean Reciprocal Rank (MRR) is the second evaluation metric~\cite{Manning2008}.
We compute MAP and MRR over a set of queries $Q$, whereby $Q$ is equivalent to the seed decisions with $|Q_{\text{WS}}|=\wsCount$ available in \ws and $|Q_{\text{OCB}}|=\ocbCount$ for \ocb.
In addition to the accuracy-oriented metrics, we evaluate the coverage and Jaccard index of the recommendations.
The coverage for the method $a$ is defined as in Equation~\ref{eq:coverage} where $D$ denotes the set of all available documents in the corpus and $D_a$ denotes the recommended documents by $a$ \cite{Ge2010}.

\begin{equation} 
\label{eq:coverage}
Cov(a)=\frac{|D_a|}{|D|}
\end{equation}

We define the Jaccard index \cite{Jaccard1912} for the similarity and diversity of two recommendation sets $R_a$ and $R_b$ from methods $a$ and $b$ for the seed $d_s$ in Equation~\ref{eq:jaccard}:

\begin{equation} 
\label{eq:jaccard}
J(a,b) = {{|R_a \cap R_b|}\over{|R_a \cup R_b|}} 
\end{equation}

We divide the evaluated methods into three categories: Word vector-, Transformer-, and citation-based methods.

\subsubsection{TF-IDF Baseline}

As a baseline method, we use the sparse document vectors from TF-IDF~\cite{Salton1975}, which are commonly used in related works~\cite{Nanda2019,Kumar2011}\footnote{We use the TF-IDF implementation from the scikit-learn framework  \cite{scikit-learn}.}.

\subsubsection{Word vector-based Methods}
The following methods are derived from word vectors, i.e., context-free word representations. %
\textbf{Paragraph Vectors}~\cite{Le2014} extend the idea of word2vec~\cite{Mikolov2013} to learning embeddings for word sequences of arbitrary length.
Paragraph Vectors using distributed bag-of-words (dbow) performed well in text similarity tasks applied on legal documents~\cite{Ash2018,Mandal2017} and other domains~\cite{Lau16}.
We train Paragraph Vectors' dbow model to 
generate document vectors for each court decision.
Like word2vec, \textbf{\gloVe}~\cite{Pennington2014} and \textbf{\fastText}~\cite{Bojanowski2017,Joulin2017} produce dense word vectors but they do not provide document vectors.
To embed a court decision as a vector, %
we compute the weighted average over its word vectors, $\vec{w_i}$, whereby the number of occurrences of the word $i$ in $d$ defines the weight $c_i$. 
Averaging of word vectors is computationally effective and yields good results for representing even longer documents \cite{Arora2017}.
For our experiments, we use word vectors made available by the corresponding authors and custom word vectors.
While GloVe vectors are pretrained on Wikipedia and Gigaword~\cite{Pennington2014}, 
fastText is pretrained on Wikipedia, UMBC webbase corpus and statmt.org news dataset~\cite{Bojanowski2017}.
Additionally, we use custom word vectors\footnote{The legal word vectors can be downloaded from our GitHub repository.} for both methods (namely \textbf{\fastTextLegal} and \textbf{\gloVeLegal}) pretrained on the joint court decision corpus extracted from \ocb and \ws 
(see Section~\ref{ssec:dataset}).
Using word vectors pretrained on different corpora, allows the evaluation of the method's cross-domain applicability.

\subsubsection{Transformer-based Methods}

As the second method category, we employ language models for deep contextual text representations based on the Transformer architecture~\cite{Vaswani2017}, namely \textbf{BERT}~\cite{Devlin2019}, \textbf{RoBERTa}~\cite{Liu2019}, Sentence Transformers (\textbf{Sentence-BERT} and \textbf{Sentence-RoBERTa})~\cite{Reimers2019}, \textbf{LongFormer}~\cite{Beltagy2020} and variations of them.
In contrast to Paragraph Vectors and average word vectors, which neglect the word order, the Transformers incorporate word positions making the text representations context-dependent.
BERT significantly improved the state-of-the-art for many NLP tasks. 
In general, BERT models are pretrained on large text corpora in an unsupervised fashion to then be fine-tuned for specific tasks like document classification~\cite{Ostendorff2019}. 
We use four variations of BERT. 
The original BERT \cite{Devlin2019} as base and large version (pretrained on Wikipedia and BookCorpus) and two BERT-base models pretrained on legal corpora. %
\textbf{Legal-JHU-BERT-base} from \citet{Holzenberger2020} which is a BERT base model but fine-tuned on the CAP corpus.
Similarly, \textbf{Legal-AUEB-BERT-base} from \citet{Chalkidis2020} is as well fine-tuned on the CAP corpus but also on other corpora (court cases and legislation from the US and EU, and US contracts).
RoBERTa improves BERT with longer training, larger batches, and removal of the next sentence prediction task for pretraining.
Sentence Transformers are fine-tuned BERT and RoBERTa models in a Siamese setting~\cite{Bromley1993} to derive semantically meaningful sentence embeddings that can be compared using cosine similarity (Sentence-BERT and Sentence-RoBERTa). 
The provided Sentence Transformers variations are \textit{nli-} or \textit{stsb}-version that are either fine-tuned on the SNLI and MNLI dataset~\cite{Bowman2015,Williams2018} or fine-tuned on the STS benchmark~\cite{Cer2017}. 
As the self-attention mechanism scales quadratically with the sequence length, the Transfomer-based methods (BERT, RoBERTa and Sentence Transformers) bound their representation to 512 tokens.
Longformer includes an attention mechanism that scales linearly with sequence length, which allows to process longer documents.
We use pretrained Longformer models as provided by \citet{Beltagy2020} and limited to 4096 tokens. 
All Transformer models apply mean-pooling to derive document vectors.
We experimented with other pooling strategies but they yield significantly lower results.
These findings agree with \citet{Reimers2019}. %
We investigate each Transformer in two variations depending on their availability and w.r.t. model size and document vector size (base with $s=768$ and large with $s=1024$).

\subsubsection{Citation-based Methods}

We explore citation-based graph methods in which documents are nodes and edges correspond to citations to generate document vectors.
Like text-based representations, citation graph embeddings have the vector size $\vec{d} \in \R^{300}$. 
With \textbf{DeepWalk}, Perozzi et al.~\cite{Perozzi2014} were the first to borrow word2vec's idea and applied it to graph network embeddings.
DeepWalk performs truncated random walks on a graph and the node embeddings are learned through the node context information encoded in these short random walks similar to the context sliding window in word2vec. 
\textbf{Walklets}~\cite{Perozzi2017} explicitly encodes multi-scale node relationships to capture community structures with the graph embedding.
Walklets generates these multi-scale relationships by sub-sampling short random walks on the graph nodes.
\textbf{BoostNE}~\cite{Li2019} is a matrix factorization-based embedding technique combined with gradient boosting. 
In \cite{Li2019}, BoostNE is applied on a citation graph from scientific papers and outperforms other graph embeddings such as DeepWalk.
Hence, we expect comparable results for the legal citation graph.
Nickel and Kiela~\cite{Nickel2017} introduced \textbf{\poincare} embeddings as a method to learn embedding in the hyperbolic space of the Poincar\'{e} ball model rather than the Euclidean space used in the aforementioned methods.
Embeddings produced in hyperbolic space are naturally equipped to model hierarchical structures~\cite{Krioukov2010}. 
Such structures can also be found in the legal citation graph in the form of different topics or jurisdictions.
For DeepWalk, Walklets, BoostNe, we use the Karate Club implementation~\cite{Rozemberczki2020}.

\subsubsection{Variations \& Hybrid Methods}

Given the conceptional differences in the evaluated methods, each method has its strength and weakness.
For further insights on these differences, we evaluate all methods with \textbf{limited text}, \textbf{vector concatenation}, and \textbf{score summation}:
Unlike the Transformers, the word vector-based methods have no maximum of input tokens.
Whether an artificial limitation of the document length improves or decreases the results is unclear.
Longer documents might add additional noise to the representation and could lead to worse results \cite{Schwarzer2016}.
To make these two method categories comparable, we include additional variations of the word vector-based methods that are limited to the first 512 or 4096 tokens of the document. %
For instance, the method \textit{\fastTextLegal (512)} has only access to the first 512 tokens.

Additionally, we explore hybrid methods that utilize text and citation information. 
Each of the single methods above yields a vector representation $\vec{d}$ for a given document $d$. 
We combine methods by concatenating their vectors.
For example, the vectors from \fastText $\vec{d}_\text{\fastText}$ and \poincare $\vec{d}_\text{\poincare}$ can be concatenated as in Equation~\ref{eq:concat}:

\begin{equation} 
\label{eq:concat}
\vec{d}=\vec{d}_\text{\fastText}||\vec{d}_\text{\poincare}
\end{equation}

The resulting vector size is the sum of the concatenated vector sizes, e.g., $s=300+300=600$.
Recommendations based on the concatenated methods are retrieved in the same fashion as the other methods, with cosine similarity.
Moreover, we combine methods by adding up their cosine similarities~\cite{Wang2016}.
The combined score of two methods is the sum of the individual scores, e.g., for method \textit{X} and method \textit{Y} the similarity of two documents $d_a$ and $d_b$ is computed as in Equation~\ref{eq:sum}. 
Methods with score summation are denoted with $X + Y$, e.g., \poincareSumFastTextLegal.

\begin{equation} 
\label{eq:sum}
sim(\vec{d_a}, \vec{d_b})=sim(\vec{d}_{\text{X}_a}, \vec{d}_{\text{X}_b})+sim(\vec{d}_{\text{Y}_a}, \vec{d}_{\text{Y}_b})
\end{equation}

Lastly, we integrate citation information into Sentence Transformers analog to the fine-tuning procedure proposed by \citet{Reimers2019}.
Based on the citation graph, we construct a dataset of positive and negative document pairs.
Two documents $d_a,d_b$ are considered as positive samples when they are connected through a citation.
Negative pairs are randomly sampled and do not share any citation.
\textbf{Sentence-Legal-AUEB-BERT-base} is the Sentence Tranformer model with Legal-AUEB-BERT-base as base model and trained with these citation information.

\section{Results}
\label{sec:results}

For our evaluation, we obtain a list of recommendations for each input document and method and then compute the performance measures accordingly.
We compute the average number of relevant recommendations, precision, recall, MRR, MAP, and coverage.

\subsection{Quantitative Evaluation}

\begin{table*}[ht]

\caption{\label{tab:overall_results}Overall scores for top $k=5$ recommendations from \ocb and \ws as the number of relevant documents, precision, recall, MRR, MAP and coverage for the \methodsCount methods and the vector sizes. The methods are divided into: baseline, word vector-based, Transformer-based, citation-based, and hybrid. High scores according to the exact numbers are \underline{underlined} (or \textbf{bold} for category-wise). $^{*}$ values were rounded up.}

\setlength{\tabcolsep}{3.5pt} %
\renewcommand{\arraystretch}{1.1} %

\begin{tabular}{lrrrrrrrrrrrrr}
\toprule
\textbf{Datasets $\rightarrow$}  &  {} & \multicolumn{6}{c}{\textbf{\ocb}} & \multicolumn{6}{c}{\textbf{\ws}} \\
 \cmidrule(lr){3-8}
\cmidrule(lr){9-14}
\textbf{Methods $\downarrow$} &   Size &       Rel. &      Prec. &      Recall &    MRR &    MAP & Cov. &      Rel. &      Prec. &      Recall &    MRR &    MAP & Cov. \\

\midrule
\rowcolor{Gray} 
TF-IDF                                 &  500000 &         1.60 &  0.320 &  0.032 &  0.363 &  0.020 &  0.487 &       1.59 &  0.318 &  0.026 &  0.389 &  0.015 &  0.446 \\
\midrule
Paragraph Vectors                      &     300 &         2.78 &  0.555 &  0.056 &  0.729 &  0.049 &  \textbf{0.892} &       2.39 &  0.477 &  0.036 &  0.629 &  0.030 &  \textbf{0.841} \\
\rowcolor{Gray} 
\fastText                            &     300 &         2.66 &  0.532 &  0.053 &  0.713 &  0.045 &  0.811 &       2.11 &  0.422 &  0.031 &  0.581 &  0.025 &  0.772 \\
\fastTextLegal                     &     300 &         \underline{\textbf{2.87}} &  \underline{\textbf{0.574}} &  \underline{\textbf{0.059}} &  \textbf{0.739} &  \underline{\textbf{0.050}} &  0.851 &       \textbf{2.39} &  \textbf{0.478} &  \textbf{0.037 }&  \textbf{0.631} &  \textbf{0.031} &  0.815 \\
\rowcolor{Gray} 
\fastTextLegal (512)                &     300 &         1.97 &  0.394 &  0.037 &  0.591 &  0.028 &  0.835 &       2.16 &  0.433 &  0.034 &  0.587 &  0.027 &  0.809 \\
\fastTextLegal (4096)               &     300 &         2.76 &  0.552 &  0.054 &  0.727 &  0.045 &  0.867 &       2.33 &  0.466 &  0.035 &  0.620 &  0.029 &  0.817 \\
\rowcolor{Gray} 
\gloVe                               &     300 &         2.68 &  0.536 &  0.054 &  0.702 &  0.046 &  0.814 &       2.06 &  0.412 &  0.033 &  0.577 &  0.026 &  0.789 \\
\gloVeLegal                        &     300 &         2.82 &  0.564 &  0.057 &  0.724 &  0.048 &  0.834 &       2.31 &  0.461 &  0.037 &  0.621 &  0.030 &  0.804 \\
\midrule

\rowcolor{Gray} 
BERT-base                   &     768 &         1.26 &  0.253 &  0.021 &  0.428 &  0.015 &  0.815 &       1.62 &  0.323 &  0.021 &  0.485 &  0.015 &  0.784 \\
BERT-large                  &    1024 &         1.35 &  0.270 &  0.022 &  0.443 &  0.016 &  0.841 &       1.82 &  0.364 &  0.023 &  0.530 &  0.018 &  0.794 \\
\rowcolor{Gray} 
Legal-JHU-BERT-base                        &     768 &         1.47 &  0.295 &  0.025 &  0.482 &  0.018 &  0.848 &       1.85 &  0.371 &  0.027 &  0.537 &  0.020 &  0.796 \\
Legal-AUEB-BERT-base      &     768 &         1.66 &  0.331 &  0.028 &  0.506 &  0.021 &  0.884 &       \textbf{2.01} &  \textbf{0.401 }&  \textbf{0.027} & \textbf{ 0.573 }& \textbf{ 0.022} &  0.813 \\
\rowcolor{Gray} 
Longformer-base              &     768 &         1.91 &  0.382 &  0.033 &  0.572 &  0.026 &  \textbf{0.892} &       1.65 &  0.329 &  0.020 &  0.514 &  0.016 &  \textbf{0.841} \\
Longformer-large             &    1024 &         \textbf{2.09} &  \textbf{0.419} &  \textbf{0.039} &  \textbf{0.614} &  \textbf{0.031} &  0.885 &       1.80 &  0.360 &  0.023 &  0.535 &  0.018 &  0.826 \\
\rowcolor{Gray} 
RoBERTa-large                     &    1024 &         1.52 &  0.305 &  0.026 &  0.481 &  0.019 &  0.843 &       1.93 &  0.387 &  0.026 &  0.553 &  0.020 &  0.782 \\
Sentence-BERT-large-nli            &    1024 &         1.03 &  0.206 &  0.018 &  0.352 &  0.013 &  0.872 &       1.37 &  0.273 &  0.017 &  0.443 &  0.012 &  0.782 \\
\rowcolor{Gray} 
Sentence-BERT-large-nli-stsb       &    1024 &         0.98 &  0.196 &  0.018 &  0.338 &  0.013 &  0.848 &       1.36 &  0.272 &  0.015 &  0.434 &  0.011 &  0.777 \\
Sentence-RoBERTa-large-nli         &    1024 &         0.92 &  0.183 &  0.016 &  0.321 &  0.011 &  0.884 &       1.18 &  0.236 &  0.013 &  0.409 &  0.009 &  0.795 \\
\midrule

\rowcolor{Gray} 
BoostNE                                &     300 &         1.29 &  0.258 &  0.022 &  0.442 &  0.016 &  0.800 &       1.24 &  0.248 &  0.016 &  0.398 &  0.013 &  0.832 \\
DeepWalk                               &     300 &         1.34 &  0.267 &  0.028 &  0.473 &  0.021 &  0.818 &       1.82 &  0.364 &  0.030 &  0.533 &  0.025 &  \underline{\textbf{0.856}} \\
\rowcolor{Gray} 
Poincaré                               &     300 &         \textbf{2.24} &  0.447 &  \textbf{0.044} &  0.629 &  \textbf{0.036} &  \underline{\textbf{0.930}} &       2.33 &  0.465 &  0.038 &  0.598 &  \textbf{0.031 }&  0.837 \\
Walklets                               &     300 &         \textbf{2.24} &  \textbf{0.448} &  0.043 &  \textbf{0.636} &  0.035 &  0.816 &       \textbf{2.35} &  \textbf{0.470} & \textbf{ 0.038} & \textbf{ 0.611} &  0.031 &  0.826 \\

\midrule

\rowcolor{Gray} 
\poincareCatFastTextLegal           &     600 &         2.36 &  0.473 &  0.048 &  0.656 &  0.041 &  0.737 &       \underline{\textbf{2.52}} &  \underline{\textbf{0.505}} &  \underline{\textbf{0.041}} &  0.638 & \underline{\textbf{0.035}} &  0.818 \\
\longformerCatFastTextLegal   &    1324 &         2.26 &  0.451 &  0.043 &  0.642 &  0.035 &  0.876 &       1.91 &  0.383 &  0.025 &  0.547 &  0.020 &  0.829 \\

\rowcolor{Gray} 
\poincareSumFastTextLegal            &     \Gape[0pt][2pt]{\makecell[r]{300\\300}} &         \textbf{2.85} &  \textbf{0.571} & \textbf{ 0.058} &  \underline{\textbf{0.746 }}&  \textbf{0.050} &  0.860 &       2.48 &  0.497 &  0.040 &  \underline{\textbf{0.646 }}&  0.034 &  0.835 \\

\poincareSumLongformer    &    \Gape[0pt][2pt]{\makecell[r]{300\\1024}} &         2.09 &  0.419 &  0.039 &  0.630 &  0.033 &  0.885 &       1.80 &  0.360 &  0.023 &  0.548 &  0.019 &  0.826 \\
\rowcolor{Gray} 
Sentence-Legal-AUEB-BERT-base &     768 &         2.19 &  0.438 &  0.039 &  0.603 &  0.031 &  \textbf{0.917} &       2.36 &  0.471 &  0.038 &  0.602 &  0.032 &  \textbf{0.849} \\
\bottomrule
\end{tabular}

\end{table*}

\subsubsection{Overall Results}

Table~\ref{tab:overall_results} presents the overall evaluation metrics for \methodsCount methods and the two datasets.
From the non-hybrid methods, \fastTextLegal yields with $0.05$ the highest MAP score on \ocb, whereas on \ws, \fastTextLegal, \poincare, and Walklets all achieve the highest MAP score of $0.031$.
The hybrid method of \poincareCatFastTextLegal outperforms the non-hybrids for \ws with $0.035$ MAP.
For \ocb, the MAP of \poincareSumFastTextLegal and \fastTextLegal are equally high.

Due to space constraints, we remove \excludedMethodsCount methods from Table~\ref{tab:overall_results} (excluded methods are in the supplementary materials\footref{fn:github}). 
From the word vector-based methods, we discard the 512 and 4096 tokens variations of Paragraph Vectors, \gloVe and \gloVeLegal, as they show a similar performance deterioration as \fastTextLegal.
The base versions of some Transformers are also excluded in favour of the better performing large versions.
Similarly, the \textit{nli} version always outperform the \textit{stsb} version of Sentence Transformers (sBERT and sRoBERTa). 
For the hybrid variations, we show only the best methods.
We also tested Node2Vec~\cite{Grover2016} and but exclude it given its low MAP scores.

Regarding the word vector-based methods, we see that the methods which are trained on the legal corpus (Paragraph Vectors, \fastTextLegal, \gloVeLegal) perform similarly well with a minor advantage by \fastTextLegal.
Moreover, there is a margin between the generic and legal word vectors even though the legal word vectors are trained on a small corpus compared to ones from the generic vectors.
The advantage of Paragraph Vectors over TF-IDF is consistent with the results from \citet{Mandal2017}.
Limiting the document length to 512 or 4096 decreases the effectiveness of \fastTextLegal.
A limit of 512 tokens decreases the MAP score to 59\% compared to all tokens on \ocb.
With 4096 tokens, the performance decline is only minor (90\% compared to all tokens).
The token limitation effect is also larger on \ocb than \ws.
The 4096 tokens version of \fastTextLegal even outperforms all Transformer methods.

Longformer-large is the best Transformer for \ocb with $0.031$ MAP.
For \ws, \legalAUEB achieves the highest MAP of 0.022, closely followed by \legalJHU.
The Longformer's theoretical advantage of processing 4096 instead of 512 tokens does not lead to better results for \ws, for which even BERT scores the same MAP of 0.018.
We generally observe that large models outperform their base counterparts\footnote{\legalJHU and \legalAUEB are only available as base version.}.
Likewise, RoBERTa has higher scores than BERT as \citet{Liu2019} suggested. 
From the Transformers category, Sentence Transformers yield the worst results. 
We assume that fine-tuning on the similarity datasets like NLI or STSB does not increase the performance since the models do not generalize well to other domains.
However, the language model fine-tuning from \legalJHU and \legalAUEB does improve the performance, whereby \legalAUEB generally outperforms \legalJHU.
For \ocb, \legalAUEB is the best model in the Transformer category in terms of MAP even though it is only used as base version.

\poincare and Walklets are by far the best methods in the citation category.
For \ws, the two citation-based methods, score the same MAP of 0.031 as \fastTextLegal.
Compared to the word vector-based methods, the citation methods do better on \ws than on \ocb.

In the category of hybrid methods, the combination of text and citations improves the performance.
For \ocb, the score summation \poincareSumFastTextLegal has the same MAP of 0.05 as \fastTextLegal but a higher MRR of 0.746.
The MRR of \poincareSumFastTextLegal is even higher than the MRR of its sub-methods \poincare (0.629) and \fastTextLegal (0.739) individually.
The concatenation of \poincareCatFastTextLegal is with 0.035 MAP the best method on \ws.
Using citation as training signal as in Sentence-Legal-AUEB-BERT also improves the performance but not as much as concatenation or summation. 
When comparing the three hybrid variations, score summation achieves overall the best results.
In the case of \ws, the concatenation's scores are below its sub-methods, while summation has at least the best sub-methods score.
Moreover, combining two text-based methods such as Longformer-large and \fastTextLegal never improves its sub-methods.

\subsubsection{Document Length}

\begin{figure*}[ht]
\includegraphics[width=\textwidth,trim=0cm 0.75cm 0cm 0cm,clip]{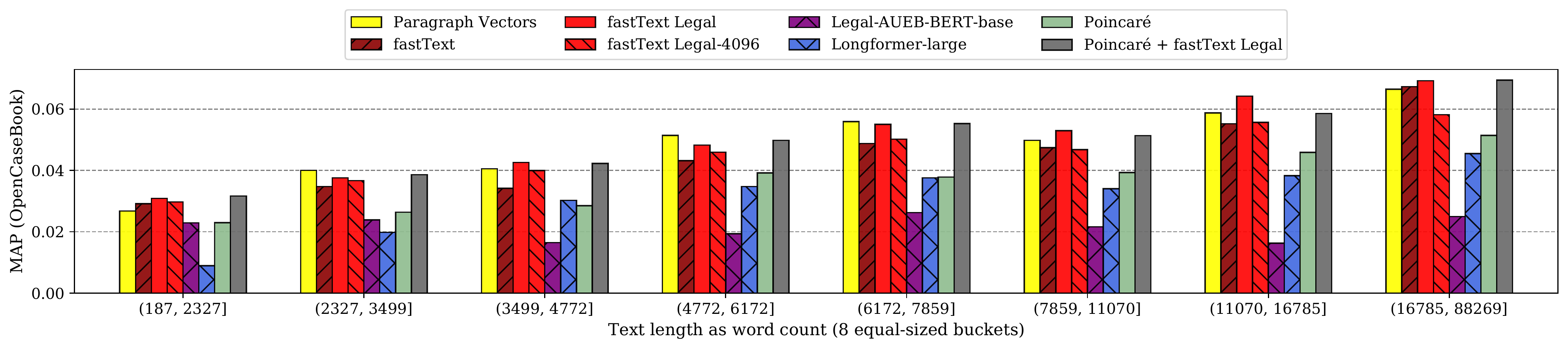}
\includegraphics[width=\textwidth,trim=0cm 0cm 0cm 1.55cm,clip]{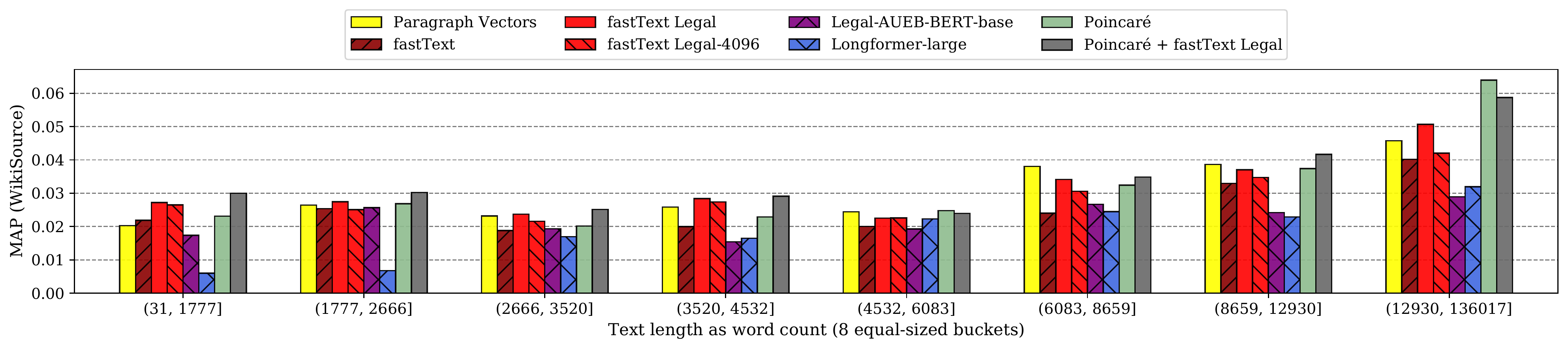}
  \caption{\label{fig:map_by_word_count}MAP wrt. words in the seed document of \ocb (top) and \ws (bottom). %
  The more words, the better the results, no peak at medium length. 
  \fastTextLegal outperforms Legal-BERT and Longformer for short documents.}
\end{figure*}

The effect of the document length on the performance in terms of MAP is displayed in Figure~\ref{fig:map_by_word_count}. 
We group the seed documents into eight equal-sized buckets (each bucket represents the equal number of  documents) depending on the word count in the document text to make the two datasets comparable.

Both datasets, \ocb and \ws, present a similar outcome. 
The MAP increases as the word count increases.
Table~\ref{tab:overall_results} presents the average overall documents and, therefore, the overall best method is not equal to the best method in some subsets.
For instance, Paragraph Vectors achieve the best results for several buckets, e.g., 4772-6172 words in \ocb or 6083-8659 words in \ws. 
The text limitation of \fastTextLegal (4096 tokens) in comparison to \fastText is also clearly visible.
The performance difference between the two methods increases as the document length increases.
For the first buckets with less than 4096 words, e.g., 187-2327 words in \ocb, one could expect no difference since the limitation does not affect the seed documents in these buckets.
However, we observe a difference since target documents are not grouped into the same buckets. %
Remarkable is that the performance difference for very long documents is less substantial. %
When comparing Longformer-large and \legalAUEB, we also see an opposing performance shift with changing word count.
While \legalAUEB's scores are relatively stable throughout all buckets, Longformer depends more on the document length. %
On the one hand, Longformer performs worse than \legalAUEB for short documents, i.e., 187-2327 words in \ocb, and 31-1777 words in \ws.
On the other hand, for documents with more words, Longformer mostly outperforms \legalAUEB by a large margin.
The citation-based method \poincare is as well affected by the document length. 
However, this effect is due to a positive correlation between word count and citation count.

\subsubsection{Citation Count}

\begin{figure*}[ht]
\includegraphics[width=\textwidth,trim=0cm 0.75cm 0cm 0cm,clip]{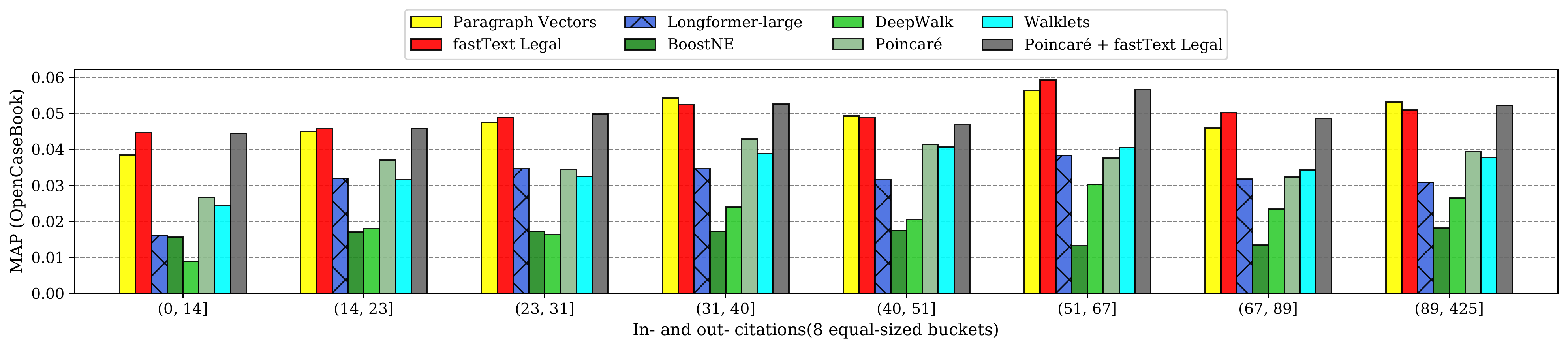}
\includegraphics[width=\textwidth,trim=0cm 0cm 0cm 1.55cm,clip]{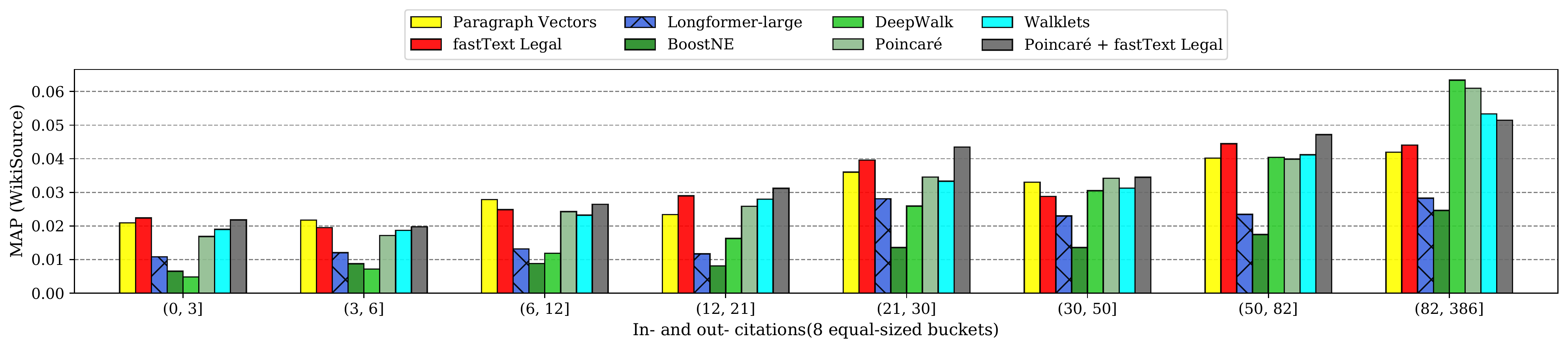}
  \caption{\label{fig:map_by_cits}MAP scores wrt. citation count for \ocb (top) and \ws (bottom). Among citation-based methods, \poincare and Walklets perform on average the best, while DeepWalk outperforms them only for \ws and when more than 82 citations are available (rightmost bucket).}
\end{figure*}

Figure~\ref{fig:map_by_cits} shows the effect of the number of in- and out-citations (i.e., edges in the citation graph) on the MAP score.
The citation analysis for \ws confirms the word count analysis.
More data leads to better results.
Instead, for \ocb, the performance of the citation-based methods peak for 31-51 citations and even decrease at 67-89 citations.
When comparing \poincare and Walklets there is no superior method and no dependency pattern is visible. 
The performance effect on DeepWalk is more substantial.
The number of citations must be above a certain threshold to allow DeepWalk to achieve competitive results.
For \ocb, the threshold is at 51-67 citations, and for \ws, it is at 30-50 citations.
Figure~\ref{fig:map_by_cits} also shows the on average higher MAP of \poincareSumFastTextLegal in comparison to the other approaches.
Citation-based methods require citations to work, whereas text methods do not have this limitation (see 0-14 citations for \ocb). 
When no citations are available, citation-based methods cannot recommend any documents, whereas the text methods still work (see 0-14 citations for \ocb). 

Our citation-based methods use only a fraction of original citation data, 70,865 citations in \ocb, and 331,498 citations in \ws, because of limitation to the documents available in the silver standards. %
For comparison, the most-cited decision from CourtListener (the underlying corpus of \ws) has 88,940 citations, whereas in experimental data of \ws the maximum number of in- and out-citations is 386.
As a result, we expect the citation-based methods, especially DeepWalk, to work even better when applied on the full corpus.

\subsubsection{Coverage and Similarity of Recommendations}

\begin{figure}
\includegraphics[width=\linewidth,trim=0cm 0.cm 0cm 0cm,clip]{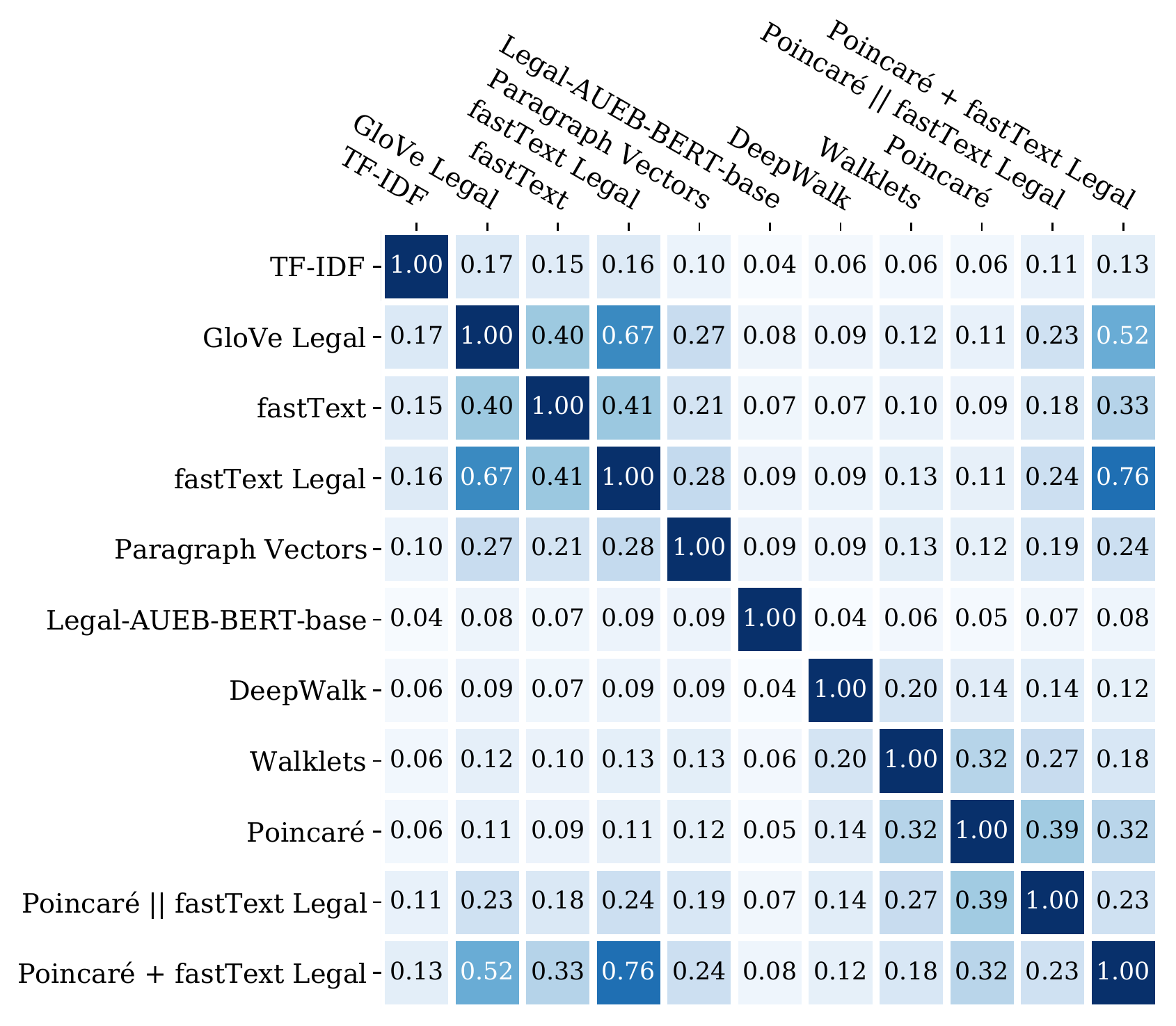}
  \caption{\label{fig:jaccard}Jaccard index for similarity or diversity of two recommendation sets (average over all seeds from the two datasets). }
\end{figure}

In addition to the accuracy-oriented metrics, Table~\ref{tab:overall_results} reports also the coverage of the recommendation methods.
A recommender systems for an expert audience should not focus on small set of most-popular items but rather provide a high coverage of the whole item collection.
However, coverage alone does not account for relevancy and, therefore, it must be contextualized with other metrics, e.g., MAP.

Overall, two citation-based methods yield the highest coverage for both datasets, i.e., \poincare for \ocb and DeepWalk for \ws.
In particular, \poincare has not only a high coverage but also high MAP scores.
Yet, the numbers do not indicate that citation-based methods have generally a higher coverage since the text-based Paragraph Vectors or Longformer-base also achieve a considerably high coverage.
The lowest coverage has by far the TF-IDF baseline.
Notable, the hybrid methods with concatenation and summation have a different effect on the coverage as on the accuracy metrics.
While the hybrid methods generally yield a higher MAP, their coverage is lower compared to their sub-methods.
Only, the Sentence-Legal-AUEB-BERT-base yields a higher coverage compared to Legal-AUEB-BERT-base.

Besides the coverage, we also analyze the similarity or diversity of the recommendations between two methods.
Figure~\ref{fig:jaccard} shows the similarity measured as Jaccard index for selected methods.
Method pairs with $J(a,b)=1$ have identical recommendations, whereas $J(a,b)=0$ means no common recommendations.
Generally speaking, the similarity of all method pairs is considerably low ($J<0.8$).
The highest similarity can be found between a hybrid method and one of its sub-methods, e.g., \poincareSumFastTextLegal and \fastTextLegal with $J=0.76$.
Apart from that, substantial similarity can be only found between pairs from the same category.
For example, the pair of the two text-based methods of \gloVeLegal and \fastTextLegal yields $J=0.67$.
Citation-based methods tend to have a lower similarity compared to the text-based methods, whereby the highest Jaccard index between two citation-based methods is achieved for Walklets and \poincare with $J=0.32$.
Like the coverage metric, the Jaccard index should be considered in relation to the accuracy results.
\gloVeLegal and \fastTextLegal yield equally high MAP scores, while having also a high recommendation's similarity.
In contrast, the MAP for \ws from \fastTextLegal and \poincare is equally high, too. 
However, their recommendation's similarity is low $J=0.11$.
Consequently, \fastTextLegal and \poincare provide relevant recommendations that are diverse from each other.
This explains the good performance of their hybrid combination.

\subsection{Qualitative Evaluation}

Due to lack of openly available gold standards, we conduct our quantitative analysis using silver standards.
Thus, we additionally conduct a qualitative evaluation with domain experts to estimate the quality of our silver standards.

\begin{table*}[ht]

\caption{Examples from \fastTextLegal and \poincare (other methods are in the supplementary material) for \textit{Mugler v. Kansas}  with relevance annotations by the silver standards (S) and domain expert (D).}\label{tab:example} 

\centering
\renewcommand{\arraystretch}{1.2} %

\begin{tabular}{crlccc|lccc}

\toprule
 {} & {} & \multicolumn{4}{c}{\textbf{\ocb}}   &  \multicolumn{4}{c}{\textbf{\ws}}  \\
\cmidrule(lr){3-6}
\cmidrule(lr){7-10}

& \multicolumn{1}{c}{\#} & \multicolumn{1}{l}{Recommendations} & \multicolumn{1}{c}{Year} & \multicolumn{1}{c}{S} & \multicolumn{1}{c}{D} & \multicolumn{1}{l}{Recommendations} & \multicolumn{1}{c}{Year} & \multicolumn{1}{c}{S} & \multicolumn{1}{c}{D} \\

\midrule
\rowcolor{Gray} 
\cellcolor{white}{} & 1 & Yick Wo v. Hopkins  & 1886 & N  & N &  Kidd v. Pearson  & 1888 & N  & Y \\
 
\cellcolor{white}{} & 2 & Munn v. Illinois  & 1876 & Y  & Y &  Lawton v. Steele  & 1894 & N  & Y \\
\rowcolor{Gray} 
\cellcolor{white}{} & 3 & LS. Dealers’ \& Butchers’  v. Crescent City LS.   & 1870 & N  & Y &  Yick Wo v. Hopkins  & 1886 & N  & N \\
 
\cellcolor{white}{} & 4 & Butchers’ Benevolent  v. Crescent City LS.   & 1872 & Y  & Y &  Geer v. Connecticut  & 1896 & N  & Y \\
\rowcolor{Gray}
\parbox[t]{2mm}{\multirow{-5}{*}{\cellcolor{white}{\rotatebox[origin=c]{90}{\fastTextLegal}}}}

\cellcolor{white}{} & 5 & Lochner v. New York  & 1905 & Y  & Y &  Groves v. Slaughter  & 1841 & Y  & N \\

\midrule
\cellcolor{white}{} & 1 & Yick Wo v. Hopkins  & 1886 & N  & N &  Rast v. Van Deman \& Lewis Co.  & 1916 & Y  & N \\

\rowcolor{Gray} 
\cellcolor{white}{} & 2 & Allgeyer v. Louisiana  & 1897 & Y  & Y &  County of Mobile v. Kimball  & 1881 & N  & N \\

\cellcolor{white}{} & 3 & Calder v. Wife  & 1798 & N  & N &  Brass v. North Dakota Ex Rel. Stoeser  & 1894 & Y  & Y \\
 \rowcolor{Gray} 
\cellcolor{white}{} & 4 & Davidson v. New Orleans  & 1877 & Y  & Y &  Erie R. Co. v. Williams  & 1914 & Y  & Y \\

\parbox[t]{2mm}{\multirow{-5}{*}{\cellcolor{white}{\rotatebox[origin=c]{90}{\poincare}}}}

\cellcolor{white}{} & 5 & Muller v. Oregon  & 1908 & Y  & Y &  Hall v. Geiger-Jones Co.  & 1917 & Y  & Y \\

\bottomrule

\end{tabular}

\end{table*}

Table~\ref{tab:example} lists one of the randomly chosen seed decisions (Mugler vs. Kansas\footnote{\url{https://www.courtlistener.com/opinion/92076/mugler-v-kansas/}}), and five recommended similar decisions, each from \fastTextLegal and \poincare. 
In Mugler vs. Kansas (1887), the court held that Kansas could constitutionally outlaw liquor sales with constitutional issues raised on substantive due process (Fourteenth Amendment) and takings (Fifth Amendment).
We provide a detail description of the cases and their relevance annotations in \extended{the supplemental material\footref{fn:github}}{Appendix~\ref{asec:manual}}.

The sample verification indicates the overall usefulness of both text-based and citation-based methods and does not contradict our quantitative findings. 
Each of the identified cases have a legal important connection to the seed case (either the Fourteenth Amendment or Fifth Amendment), although it is difficult to say whether the higher-ranked cases are more similar along an important topical dimension. 
The rankings do not appear to be driven by facts presented in the case as most of them have not to do with alcohol bans.
Only Kidd vs. Pearson (1888) is about liquor sales as the seed decision.
The samples also do not reveal considerable differences between text- and citation-based similarity.
Moreover, we cannot confirm the findings from \citet{Schwarzer2016}, which suggests that text-based methods are focused on specific terms and citation yield mostly broadly related recommendations.
With regards to the silver standards, the domain expert annotations agree in 14 of 20 cases (70\%).
In only two cases the domain expert classify a recommendation as irrelevant despite being classified as relevant in the silver standard.

\section{Discussion}
\label{sec:discussion}

Our experiments explore the applicability of the latest advances in research to the use case of legal literature recommendations.
Existing studies on legal recommendations typically rely on small-scale user studies and are therefore limited in the number of approaches that they can evaluate (Section~\ref{sec:related}).
For this study, we utilize relevance annotations from two publicly available sources, i.e., \ocb and \ws.
These annotations does not only enable us to evaluate the recommendations of \silverCountText documents but also the comparison of in total \allMethodsCount methods and their variations of which \methodsCount methods are presented in this paper.

Our extensive evaluation shows a large variance in the recommendation performance.
Such a variance is known from other studies~\cite{Beel2016}.
There is no single method that yields the highest scores across all metrics and all datasets.
Despite that, \fastTextLegal is on average the best of all \allMethodsCount methods.
\fastTextLegal yields the highest MAP for \ocb, while for \ws only hybrid methods outperform \fastTextLegal.
Also, the coverage of \fastTextLegal is considerably high for both datasets.
Simultaneously, \fastTextLegal is robust to corner cases since neither very short nor very long documents reduce \fastTextLegal's performance substantially.
These results confirm the findings from \citet{Arora2017} that average word vectors are ``simple but tough-to-beat baseline''.
Regarding baselines, our TF-IDF baseline yields one of the worst results. 
In terms of accuracy metrics, only some Transformers are worse than TF-IDF, but especially TF-IDF's coverage is the lowest by a large margin.
With a coverage below 50\%, TF-IDF fails to provide diverse recommendations that are desirable for legal literature research.

The transfer of research advances to the legal domain is one aspect of our experiments.
Thus, the performance of Transformers and citation embeddings is of particular interest.
Despite the success of Transformers for many NLP tasks, Transformers yield on average the worst results for representing lengthy documents written in legal English.
The other two method categories, word vector-based, and citation-based methods, surpass Transformers.

The word vector-based methods achieve overall the best results among the non-hybrid methods.
All word vector-based methods with in-domain training, i.e., Paragraph Vectors, \fastTextLegal, and \gloVeLegal, perform similarly good with a minor advantage by \fastTextLegal.
Their similar performance aligns with the large overlap among their recommendations. %
Despite a small corpus of 65,635 documents, the in-domain training generally improves the performance as the gap between the out-of-domain \fastText and \fastTextLegal shows.
Given that the training of custom word vectors is feasible on commodity hardware, in-domain training is advised.
More significant than the gap between in- and out-of-domain word vectors is the effect of limited document lengths.
For \ocb, the \fastTextLegal variation limited to the first 512 tokens has only 52\% of the MAP of the full-text method. For \ws, the performance decline exists as well but is less significant.
This effect highlights the advantage of the word vector-based methods that they derive meaningful representations of documents with arbitrary length.

The evaluated Transformers cannot process documents of arbitrary length but are either limited to 512 or 4096 tokens.
This limitation contributes to Transformers' low performance.
For instance, Longformer-large's MAP is almost twice as high as BERT-large's MAP on \ocb.
However, for \ws both models yield the same MAP scores.
For \ws, the in-domain pretraining as a larger effect than the token limit since \legalAUEB achieves the best results among the Transformers.
Regarding the Transformer pretraining, the difference between \legalJHU and \legalAUEB shows the effect between two pretraining approaches.
The corpora and the hyperparameter settings used during pretraining are crucial.
Even though \legalJHU was exclusively pretrained on the CAP corpus, which has a high overlap with \ocb, \legalAUEB still outperforms \legalJHU on \ocb.
Given these findings, we expect the performance of Transformers could be improved by increasing the token limit beyond the 4096 tokens and by additional in-domain pretraining.
Such improvements are technically possible but add significant computational effort.
In contrast to word vectors, Transformers are not trained on commodity hardware but on GPUs.
Especially long-sequence Transformers such as the Longformer require GPUs with large memory. 
Such hardware may not be available in production deployments.
Moreover, the computational effort must be seen in relation to the other methods.
Put differently, even \fastTextLegal limited to 512 tokens outperforms all Transformers.

Concerning the citation embeddings, we consider \poincare, closely followed by Walklets, as the best method.
In particular, the two methods outperform the other citation methods for documents even when only a few citations are available, which makes them attractive for legal research.
\poincare also provides the highest coverage for \ocb, emphasizing its quality for literature recommendations.
For \ws, DeepWalk has the highest coverage despite yielding generally low accuracy scores.
As Figure~\ref{fig:map_by_cits} shows, DeepWalk's MAP score improves substantially as the number of citations increases.
Therefore, we expect that DeepWalk but also the other citation methods would perform even better when applied on larger citation graph.
The analysis of recommendation similarity also shows little overlap between the citation-based methods and the text-based methods (Figure~\ref{fig:jaccard}).
This indicates that the two approaches complement each other and motivates the use of hybrid methods.

Related work has already shown the benefit of hybrid methods for literature recommendations~\cite{Wiggers2019,Bhattacharya2020}.
Our experiments confirm these findings.
The simple approaches of score summation or vector concatenation can improve the results.
In particular, \poincareSumFastTextLegal never leads to a decline in performance.
Instead, it increases the performance for corner cases in which one of the sub-methods performs poorly.
Vector concatenation has mixed effects on the performance, e.g., positive effect for \ws and negative effect for \ocb.
Using citations as training data in Sentence Transformers can also be considered as a hybrid method that improves the performance.
However, this requires additional effort for training a new Sentence Transformer model.

As we discuss in Section~\ref{ssec:dataset}, we consider \ocb and \ws more of silver than gold standards.
With the qualitative evaluation, we mitigate the risk of misinterpreting the quantitative results, whereby we acknowledge our small sample size.
The overall agreement with the domain expert is high.
The expert tends to classify more recommendations as relevant than the silver standards, i.e., relevant recommendations are missed.
This explains the relatively low recall from the quantitative evaluation.
In a user study, we would expect only minor changes in the ranking of methods with similar scores, e.g., \fastTextLegal and \gloVeLegal.
The overall ranking among the method categories would remain the same.
The benefit of our silver standards is the number of available relevance annotations.
The number of annotations in related user studies is with up to 50 annotations rather low. %
Instead, our silver standards provide a magnitude more relevance annotations.
Almost 3,000 relevance annotations enable evaluations regarding text length, citation count, or other properties that would be otherwise magnitudes more difficult. 
Similarly, the user studies are difficult to reproduce as their data is mostly unavailable.
This leads to reproducibility being an issue in recommender system research~\cite{Beel2016}.
The open license of the silver standards allows the sharing of all evaluation data and, therefore, contributes to more reproducibility.
In summary, the proposed datasets bring great value to the field, overcoming eventual shortcomings.

\section{Conclusion} %
\label{sec:conclusions}

We present an extensive empirical evaluation of \methodsCount document representation methods in the context of legal literature recommendations.
In contrast to previous small-scale studies, we evaluate the methods over two document corpora containing \silverCountText documents (\ocbCountText from \ocb and \wsCountText from \ws).
We underpin our findings with a sample-based qualitative evaluation. 
Our analysis of the results reveals \fastTextLegal (averaged fastText word vectors trained on our corpora) as the overall best performing method.
Moreover, we find that all methods have a low overlap between their recommendations and are vulnerable to certain dataset characteristics like text length and number of citations available. 
To mitigate the weakness of single methods and to increase recommendation diversity, we propose hybrid methods like score summation of \fastTextLegal and \poincare that outperforms all other methods on both datasets.
Although there are limitations in the experimental evaluation due to the lack of openly available ground truth data, we are able to draw meaningful conclusions for the behavior of text-based and citation-based document embeddings in the context of legal document recommendation. 
Our source code, trained models, and datasets are openly available to encourage further research\footnote{\label{fn:github}GitHub repository: \url{https://github.com/malteos/legal-document-similarity}}.

\begin{acks}

We would like to thank Christoph Alt, Till Blume, and the anonymous reviewers for their comments.
The research presented in this article is funded by the German Federal Ministry of Education and Research (BMBF) through the project QURATOR\extended{}{~\cite{Rehm2020}} (Unternehmen Region, Wachstumskern, no. 03WKDA1A) and by the project LYNX\extended{}{~\cite{rehm2019c}}, which has received funding from the EU’s Horizon 2020 research and innovation program under grant agreement no. 780602.

\end{acks}

\bibliographystyle{ACM-Reference-Format}
\bibliography{references}


\begin{thebibliography}{55}


\ifx \showCODEN    \undefined \def \showCODEN     #1{\unskip}     \fi
\ifx \showDOI      \undefined \def \showDOI       #1{#1}\fi
\ifx \showISBNx    \undefined \def \showISBNx     #1{\unskip}     \fi
\ifx \showISBNxiii \undefined \def \showISBNxiii  #1{\unskip}     \fi
\ifx \showISSN     \undefined \def \showISSN      #1{\unskip}     \fi
\ifx \showLCCN     \undefined \def \showLCCN      #1{\unskip}     \fi
\ifx \shownote     \undefined \def \shownote      #1{#1}          \fi
\ifx \showarticletitle \undefined \def \showarticletitle #1{#1}   \fi
\ifx \showURL      \undefined \def \showURL       {\relax}        \fi
\providecommand\bibfield[2]{#2}
\providecommand\bibinfo[2]{#2}
\providecommand\natexlab[1]{#1}
\providecommand\showeprint[2][]{arXiv:#2}

\bibitem[\protect\citeauthoryear{Arora, Liang, and Ma}{Arora
  et~al\mbox{.}}{2017}]%
        {Arora2017}
\bibfield{author}{\bibinfo{person}{Sanjeev Arora}, \bibinfo{person}{Yingyu
  Liang}, {and} \bibinfo{person}{Tengyu Ma}.} \bibinfo{year}{2017}\natexlab{}.
\newblock \showarticletitle{{A simple but though Baseline for Sentence
  Embeddings}}. In \bibinfo{booktitle}{\emph{5th International Conference on
  Learning Representations (ICLR 2017)}}, Vol.~\bibinfo{volume}{15}.
  \bibinfo{pages}{416--424}.
\newblock


\bibitem[\protect\citeauthoryear{Ash and Chen}{Ash and Chen}{2018}]%
        {Ash2018}
\bibfield{author}{\bibinfo{person}{Elliott Ash} {and}
  \bibinfo{person}{Daniel~L. Chen}.} \bibinfo{year}{2018}\natexlab{}.
\newblock \showarticletitle{{Case Vectors: Spatial Representations of the Law
  Using Document Embeddings}}.
\newblock \bibinfo{journal}{\emph{SSRN Electronic Journal}}
  \bibinfo{volume}{11}, \bibinfo{number}{2017} (\bibinfo{date}{may}
  \bibinfo{year}{2018}), \bibinfo{pages}{313--337}.
\newblock
\showISSN{1556-5068}
\urldef\tempurl%
\url{https://doi.org/10.2139/ssrn.3204926}
\showDOI{\tempurl}


\bibitem[\protect\citeauthoryear{Bai, Wang, Lee, Yang, Kong, and Xia}{Bai
  et~al\mbox{.}}{2019}]%
        {bai2019scientific}
\bibfield{author}{\bibinfo{person}{Xiaomei Bai}, \bibinfo{person}{Mengyang
  Wang}, \bibinfo{person}{Ivan Lee}, \bibinfo{person}{Zhuo Yang},
  \bibinfo{person}{Xiangjie Kong}, {and} \bibinfo{person}{Feng Xia}.}
  \bibinfo{year}{2019}\natexlab{}.
\newblock \showarticletitle{Scientific paper recommendation: A survey}.
\newblock \bibinfo{journal}{\emph{IEEE Access}}  \bibinfo{volume}{7}
  (\bibinfo{year}{2019}), \bibinfo{pages}{9324--9339}.
\newblock


\bibitem[\protect\citeauthoryear{Beel, Breitinger, Langer, Lommatzsch, and
  Gipp}{Beel et~al\mbox{.}}{2016}]%
        {Beel2016}
\bibfield{author}{\bibinfo{person}{Joeran Beel}, \bibinfo{person}{Corinna
  Breitinger}, \bibinfo{person}{Stefan Langer}, \bibinfo{person}{Andreas
  Lommatzsch}, {and} \bibinfo{person}{Bela Gipp}.}
  \bibinfo{year}{2016}\natexlab{}.
\newblock \showarticletitle{{Towards reproducibility in recommender-systems
  research}}.
\newblock \bibinfo{journal}{\emph{User Modeling and User-Adapted Interaction
  (UMAI)}}  \bibinfo{volume}{26} (\bibinfo{year}{2016}).
\newblock


\bibitem[\protect\citeauthoryear{Beltagy, Peters, and Cohan}{Beltagy
  et~al\mbox{.}}{2020}]%
        {Beltagy2020}
\bibfield{author}{\bibinfo{person}{Iz Beltagy}, \bibinfo{person}{Matthew~E.
  Peters}, {and} \bibinfo{person}{Arman Cohan}.}
  \bibinfo{year}{2020}\natexlab{}.
\newblock \showarticletitle{{Longformer: The Long-Document Transformer}}.
\newblock  (\bibinfo{year}{2020}).
\newblock
\showeprint[arxiv]{2004.05150}


\bibitem[\protect\citeauthoryear{Bhattacharya, Ghosh, Pal, and
  Ghosh}{Bhattacharya et~al\mbox{.}}{2020}]%
        {Bhattacharya2020}
\bibfield{author}{\bibinfo{person}{Paheli Bhattacharya},
  \bibinfo{person}{Kripabandhu Ghosh}, \bibinfo{person}{Arindam Pal}, {and}
  \bibinfo{person}{Saptarshi Ghosh}.} \bibinfo{year}{2020}\natexlab{}.
\newblock \showarticletitle{{Methods for Computing Legal Document Similarity: A
  Comparative Study}}.
\newblock  (\bibinfo{year}{2020}).
\newblock
\showeprint[arxiv]{2004.12307}


\bibitem[\protect\citeauthoryear{Blei, Ng, and Jordan}{Blei
  et~al\mbox{.}}{2003}]%
        {blei2003latent}
\bibfield{author}{\bibinfo{person}{David~M Blei}, \bibinfo{person}{Andrew~Y
  Ng}, {and} \bibinfo{person}{Michael~I Jordan}.}
  \bibinfo{year}{2003}\natexlab{}.
\newblock \showarticletitle{Latent dirichlet allocation}.
\newblock \bibinfo{journal}{\emph{Journal of machine Learning research}}
  \bibinfo{volume}{3}, \bibinfo{number}{Jan} (\bibinfo{year}{2003}),
  \bibinfo{pages}{993--1022}.
\newblock


\bibitem[\protect\citeauthoryear{Boella, Caro, Humphreys, Robaldo, Rossi, and
  van~der Torre}{Boella et~al\mbox{.}}{2016}]%
        {Boella2016}
\bibfield{author}{\bibinfo{person}{Guido Boella}, \bibinfo{person}{Luigi~Di
  Caro}, \bibinfo{person}{Llio Humphreys}, \bibinfo{person}{Livio Robaldo},
  \bibinfo{person}{Piercarlo Rossi}, {and} \bibinfo{person}{Leendert van~der
  Torre}.} \bibinfo{year}{2016}\natexlab{}.
\newblock \showarticletitle{{Eunomos, a legal document and knowledge management
  system for the Web to provide relevant, reliable and up-to-date information
  on the law}}.
\newblock \bibinfo{journal}{\emph{Artificial Intelligence and Law}}
  \bibinfo{volume}{24}, \bibinfo{number}{3} (\bibinfo{year}{2016}),
  \bibinfo{pages}{245--283}.
\newblock
\showISSN{15728382}


\bibitem[\protect\citeauthoryear{Boer and Winkels}{Boer and Winkels}{2016}]%
        {Boer2016}
\bibfield{author}{\bibinfo{person}{Alexander Boer} {and}
  \bibinfo{person}{Radboud Winkels}.} \bibinfo{year}{2016}\natexlab{}.
\newblock \showarticletitle{{Making a cold start in legal recommendation: An
  experiment}}.
\newblock \bibinfo{journal}{\emph{Frontiers in Artificial Intelligence and
  Applications}}  \bibinfo{volume}{294} (\bibinfo{year}{2016}),
  \bibinfo{pages}{131--136}.
\newblock
\showISBNx{9781614997252}
\showISSN{09226389}
\urldef\tempurl%
\url{https://doi.org/10.3233/978-1-61499-726-9-131}
\showDOI{\tempurl}


\bibitem[\protect\citeauthoryear{Bojanowski, Grave, Joulin, and
  Mikolov}{Bojanowski et~al\mbox{.}}{2017}]%
        {Bojanowski2017}
\bibfield{author}{\bibinfo{person}{Piotr Bojanowski}, \bibinfo{person}{Edouard
  Grave}, \bibinfo{person}{Armand Joulin}, {and} \bibinfo{person}{Tomas
  Mikolov}.} \bibinfo{year}{2017}\natexlab{}.
\newblock \showarticletitle{{Enriching Word Vectors with Subword Information}}.
\newblock \bibinfo{journal}{\emph{Transactions of the Association for
  Computational Linguistics}}  \bibinfo{volume}{5} (\bibinfo{year}{2017}),
  \bibinfo{pages}{135--146}.
\newblock


\bibitem[\protect\citeauthoryear{Bowman, Angeli, Potts, and Manning}{Bowman
  et~al\mbox{.}}{2015}]%
        {Bowman2015}
\bibfield{author}{\bibinfo{person}{Samuel~R. Bowman}, \bibinfo{person}{Gabor
  Angeli}, \bibinfo{person}{Christopher Potts}, {and}
  \bibinfo{person}{Christopher~D. Manning}.} \bibinfo{year}{2015}\natexlab{}.
\newblock \showarticletitle{{A large annotated corpus for learning natural
  language inference}}.
\newblock \bibinfo{journal}{\emph{Proceedings of EMNLP}}
  (\bibinfo{year}{2015}), \bibinfo{pages}{632--642}.
\newblock
\showISBNx{9781941643327}


\bibitem[\protect\citeauthoryear{Bromley, Bentz, Bottou, Guyon, Lecun, Moore,
  Sackinger, and Shah}{Bromley et~al\mbox{.}}{1993}]%
        {Bromley1993}
\bibfield{author}{\bibinfo{person}{Jane Bromley}, \bibinfo{person}{J.W. Bentz},
  \bibinfo{person}{Leon Bottou}, \bibinfo{person}{I. Guyon},
  \bibinfo{person}{Yann Lecun}, \bibinfo{person}{C. Moore},
  \bibinfo{person}{Eduard Sackinger}, {and} \bibinfo{person}{R. Shah}.}
  \bibinfo{year}{1993}\natexlab{}.
\newblock \showarticletitle{{Signature verification using a Siamese time delay
  neural network}}.
\newblock \bibinfo{journal}{\emph{International Journal of Pattern Recognition
  and Artificial Intelligence}} \bibinfo{volume}{7}, \bibinfo{number}{4}
  (\bibinfo{year}{1993}).
\newblock
\showISSN{03022838}


\bibitem[\protect\citeauthoryear{Cer, Diab, Agirre, Lopez-Gazpio, and
  Specia}{Cer et~al\mbox{.}}{2017}]%
        {Cer2017}
\bibfield{author}{\bibinfo{person}{Daniel Cer}, \bibinfo{person}{Mona Diab},
  \bibinfo{person}{Eneko Agirre}, \bibinfo{person}{I{\~n}igo Lopez-Gazpio},
  {and} \bibinfo{person}{Lucia Specia}.} \bibinfo{year}{2017}\natexlab{}.
\newblock \showarticletitle{{S}em{E}val-2017 Task 1: Semantic Textual
  Similarity Multilingual and Crosslingual Focused Evaluation}. In
  \bibinfo{booktitle}{\emph{Proc. of the 11th International Workshop on
  Semantic Evaluation ({S}em{E}val-2017)}}. \bibinfo{publisher}{ACL},
  \bibinfo{address}{Vancouver, Canada}, \bibinfo{pages}{1--14}.
\newblock


\bibitem[\protect\citeauthoryear{Chalkidis, Fergadiotis, Malakasiotis, Aletras,
  and Androutsopoulos}{Chalkidis et~al\mbox{.}}{2020}]%
        {Chalkidis2020}
\bibfield{author}{\bibinfo{person}{Ilias Chalkidis}, \bibinfo{person}{Manos
  Fergadiotis}, \bibinfo{person}{Prodromos Malakasiotis},
  \bibinfo{person}{Nikolaos Aletras}, {and} \bibinfo{person}{Ion
  Androutsopoulos}.} \bibinfo{year}{2020}\natexlab{}.
\newblock \showarticletitle{{LEGAL-BERT: The Muppets straight out of Law
  School}}. In \bibinfo{booktitle}{\emph{Findings of the Association for
  Computational Linguistics: EMNLP 2020}}. \bibinfo{publisher}{ACL},
  \bibinfo{address}{Stroudsburg, PA, USA}, \bibinfo{pages}{2898--2904}.
\newblock
\showISSN{23318422}


\bibitem[\protect\citeauthoryear{Devlin, Chang, Lee, and Toutanova}{Devlin
  et~al\mbox{.}}{2019}]%
        {Devlin2019}
\bibfield{author}{\bibinfo{person}{Jacob Devlin}, \bibinfo{person}{Ming-Wei
  Chang}, \bibinfo{person}{Kenton Lee}, {and} \bibinfo{person}{Kristina
  Toutanova}.} \bibinfo{year}{2019}\natexlab{}.
\newblock \showarticletitle{{BERT: Pre-training of Deep Bidirectional
  Transformers for Language Understanding}}. In \bibinfo{booktitle}{\emph{Proc.
  of the 2019 Conf. of the NAACL}}. \bibinfo{publisher}{ACL},
  \bibinfo{address}{Minneapolis, Minnesota}, \bibinfo{pages}{4171--4186}.
\newblock


\bibitem[\protect\citeauthoryear{Ge, Delgado-Battenfeld, and Jannach}{Ge
  et~al\mbox{.}}{2010}]%
        {Ge2010}
\bibfield{author}{\bibinfo{person}{Mouzhi Ge}, \bibinfo{person}{Carla
  Delgado-Battenfeld}, {and} \bibinfo{person}{Dietmar Jannach}.}
  \bibinfo{year}{2010}\natexlab{}.
\newblock \showarticletitle{{Beyond accuracy: evaluating recommender systems by
  coverage and serendipity}}. In \bibinfo{booktitle}{\emph{Proceedings of the
  fourth ACM conference on Recommender systems - RecSys '10}}.
  \bibinfo{publisher}{ACM Press}, \bibinfo{address}{New York, New York, USA},
  \bibinfo{pages}{257}.
\newblock
\showISBNx{9781605589060}


\bibitem[\protect\citeauthoryear{Grover and Leskovec}{Grover and
  Leskovec}{2016}]%
        {Grover2016}
\bibfield{author}{\bibinfo{person}{Aditya Grover} {and} \bibinfo{person}{Jure
  Leskovec}.} \bibinfo{year}{2016}\natexlab{}.
\newblock \showarticletitle{{node2vec: Scalable Feature Learning for
  Networks}}. In \bibinfo{booktitle}{\emph{Proc. of the 22nd ACM SIGKDD Int.
  Conf. on Knowledge Discovery and Data Mining - KDD '16}}.
  \bibinfo{publisher}{ACM Press}, \bibinfo{address}{New York, New York, USA},
  \bibinfo{pages}{855--864}.
\newblock
\showISBNx{9781450342322}


\bibitem[\protect\citeauthoryear{Holzenberger, Blair-Stanek, and
  Durme}{Holzenberger et~al\mbox{.}}{2020}]%
        {Holzenberger2020}
\bibfield{author}{\bibinfo{person}{Nils Holzenberger}, \bibinfo{person}{Andrew
  Blair-Stanek}, {and} \bibinfo{person}{Benjamin~Van Durme}.}
  \bibinfo{year}{2020}\natexlab{}.
\newblock \showarticletitle{{A dataset for statutory reasoning in tax law
  entailment and question answering}}. In \bibinfo{booktitle}{\emph{Proceedings
  of the 2020 Natural Legal Language Processing Workshop}}.
  \bibinfo{pages}{31--38}.
\newblock
\showISSN{16130073}


\bibitem[\protect\citeauthoryear{Jaccard}{Jaccard}{1912}]%
        {Jaccard1912}
\bibfield{author}{\bibinfo{person}{Paul Jaccard}.}
  \bibinfo{year}{1912}\natexlab{}.
\newblock \showarticletitle{{The Distribution of the Flora in the Alpine
  Zone}}.
\newblock \bibinfo{journal}{\emph{New Phytologist}} \bibinfo{volume}{11},
  \bibinfo{number}{2} (\bibinfo{date}{feb} \bibinfo{year}{1912}),
  \bibinfo{pages}{37--50}.
\newblock
\showISSN{0028-646X}


\bibitem[\protect\citeauthoryear{Joulin, Grave, Bojanowski, and Mikolov}{Joulin
  et~al\mbox{.}}{2017}]%
        {Joulin2017}
\bibfield{author}{\bibinfo{person}{Armand Joulin}, \bibinfo{person}{Edouard
  Grave}, \bibinfo{person}{Piotr Bojanowski}, {and} \bibinfo{person}{Tomas
  Mikolov}.} \bibinfo{year}{2017}\natexlab{}.
\newblock \showarticletitle{{Bag of Tricks for Efficient Text Classification}}.
  In \bibinfo{booktitle}{\emph{Proceedings of the 15th Conference of the
  European Chapter of the Association for Computational Linguistics: Volume 2,
  Short Papers}}. \bibinfo{publisher}{ACL}, \bibinfo{address}{Stroudsburg, PA,
  USA}, \bibinfo{pages}{427--431}.
\newblock


\bibitem[\protect\citeauthoryear{Krioukov, Papadopoulos, Kitsak, Vahdat, and
  Bogu{\~{n}}{\'{a}}}{Krioukov et~al\mbox{.}}{2010}]%
        {Krioukov2010}
\bibfield{author}{\bibinfo{person}{Dmitri Krioukov},
  \bibinfo{person}{Fragkiskos Papadopoulos}, \bibinfo{person}{Maksim Kitsak},
  \bibinfo{person}{Amin Vahdat}, {and} \bibinfo{person}{Mari{\'{a}}n
  Bogu{\~{n}}{\'{a}}}.} \bibinfo{year}{2010}\natexlab{}.
\newblock \showarticletitle{{Hyperbolic geometry of complex networks}}.
\newblock \bibinfo{journal}{\emph{Physical Review E - Statistical, Nonlinear,
  and Soft Matter Physics}} \bibinfo{volume}{82}, \bibinfo{number}{3}
  (\bibinfo{year}{2010}), \bibinfo{pages}{1--18}.
\newblock
\showISSN{15393755}


\bibitem[\protect\citeauthoryear{Kumar, Reddy, Reddy, and Singh}{Kumar
  et~al\mbox{.}}{2011}]%
        {Kumar2011}
\bibfield{author}{\bibinfo{person}{Sushanta Kumar}, \bibinfo{person}{P.~Krishna
  Reddy}, \bibinfo{person}{V.~Balakista Reddy}, {and} \bibinfo{person}{Aditya
  Singh}.} \bibinfo{year}{2011}\natexlab{}.
\newblock \showarticletitle{{Similarity analysis of legal judgments}}.
\newblock \bibinfo{journal}{\emph{Compute 2011 - 4th Annual ACM Bangalore
  Conference}} (\bibinfo{year}{2011}).
\newblock
\showISBNx{9781450307505}
\urldef\tempurl%
\url{https://doi.org/10.1145/1980422.1980439}
\showDOI{\tempurl}


\bibitem[\protect\citeauthoryear{Landthaler, Waltl, Holl, and
  Matthes}{Landthaler et~al\mbox{.}}{2016}]%
        {Landthaler2016}
\bibfield{author}{\bibinfo{person}{J{\"{o}}rg Landthaler},
  \bibinfo{person}{Bernhard Waltl}, \bibinfo{person}{Patrick Holl}, {and}
  \bibinfo{person}{Florian Matthes}.} \bibinfo{year}{2016}\natexlab{}.
\newblock \showarticletitle{{Extending full text search for legal document
  collections using word embeddings}}.
\newblock \bibinfo{journal}{\emph{Frontiers in Artificial Intelligence and
  Applications}}  \bibinfo{volume}{294} (\bibinfo{year}{2016}),
  \bibinfo{pages}{73--82}.
\newblock
\showISBNx{9781614997252}
\showISSN{09226389}


\bibitem[\protect\citeauthoryear{Lastres}{Lastres}{2013}]%
        {Lastres2013}
\bibfield{author}{\bibinfo{person}{Steven~A. Lastres}.}
  \bibinfo{year}{2013}\natexlab{}.
\newblock \bibinfo{booktitle}{\emph{{Rebooting Legal Research in a Digital
  Age}}}.
\newblock \bibinfo{type}{{T}echnical {R}eport}.
  \bibinfo{institution}{LexisNexis}.
\newblock
\urldef\tempurl%
\url{https://www.lexisnexis.com/documents/pdf/20130806061418\_large.pdf}
\showURL{%
\tempurl}


\bibitem[\protect\citeauthoryear{Lau and Baldwin}{Lau and Baldwin}{2016}]%
        {Lau16}
\bibfield{author}{\bibinfo{person}{J.~H. Lau} {and} \bibinfo{person}{T.
  Baldwin}.} \bibinfo{year}{2016}\natexlab{}.
\newblock \showarticletitle{An Empirical Evaluation of doc2vec with Practical
  Insights into Document Embedding Generation}. In
  \bibinfo{booktitle}{\emph{Proceedings Workshop on Representation Learning for
  NLP}}.
\newblock
\urldef\tempurl%
\url{https://doi.org/10.18653/v1/w16-1609}
\showDOI{\tempurl}


\bibitem[\protect\citeauthoryear{Le and Mikolov}{Le and Mikolov}{2014}]%
        {Le2014}
\bibfield{author}{\bibinfo{person}{Quoc~V. Le} {and} \bibinfo{person}{Tomas
  Mikolov}.} \bibinfo{year}{2014}\natexlab{}.
\newblock \showarticletitle{{Distributed Representations of Sentences and
  Documents}}.
\newblock \bibinfo{journal}{\emph{Int. Conf. on Machine Learning}}
  \bibinfo{volume}{32} (\bibinfo{year}{2014}), \bibinfo{pages}{1188--1196}.
\newblock


\bibitem[\protect\citeauthoryear{Li, Wu, Guo, Liu, and Liu}{Li
  et~al\mbox{.}}{2019}]%
        {Li2019}
\bibfield{author}{\bibinfo{person}{Jundong Li}, \bibinfo{person}{Liang Wu},
  \bibinfo{person}{Ruocheng Guo}, \bibinfo{person}{Chenghao Liu}, {and}
  \bibinfo{person}{Huan Liu}.} \bibinfo{year}{2019}\natexlab{}.
\newblock \showarticletitle{{Multi-level network embedding with boosted
  low-rank matrix approximation}}. In \bibinfo{booktitle}{\emph{Proceedings of
  the 2019 IEEE/ACM International Conference on Advances in Social Networks
  Analysis and Mining}}. \bibinfo{publisher}{ACM}, \bibinfo{address}{New York,
  NY, USA}, \bibinfo{pages}{49--56}.
\newblock
\showISBNx{9781450368681}


\bibitem[\protect\citeauthoryear{Liu, Ott, Goyal, Du, Joshi, Chen, Levy, Lewis,
  Zettlemoyer, and Stoyanov}{Liu et~al\mbox{.}}{2019}]%
        {Liu2019}
\bibfield{author}{\bibinfo{person}{Yinhan Liu}, \bibinfo{person}{Myle Ott},
  \bibinfo{person}{Naman Goyal}, \bibinfo{person}{Jingfei Du},
  \bibinfo{person}{Mandar Joshi}, \bibinfo{person}{Danqi Chen},
  \bibinfo{person}{Omer Levy}, \bibinfo{person}{Mike Lewis},
  \bibinfo{person}{Luke Zettlemoyer}, {and} \bibinfo{person}{Veselin
  Stoyanov}.} \bibinfo{year}{2019}\natexlab{}.
\newblock \showarticletitle{{RoBERTa: A Robustly Optimized BERT Pretraining
  Approach}}.
\newblock  (\bibinfo{year}{2019}).
\newblock
\showeprint[arxiv]{1907.11692}


\bibitem[\protect\citeauthoryear{Mandal, Chaki, Saha, Ghosh, Pal, and
  Ghosh}{Mandal et~al\mbox{.}}{2017}]%
        {Mandal2017}
\bibfield{author}{\bibinfo{person}{Arpan Mandal}, \bibinfo{person}{Raktim
  Chaki}, \bibinfo{person}{Sarbajit Saha}, \bibinfo{person}{Kripabandhu Ghosh},
  \bibinfo{person}{Arindam Pal}, {and} \bibinfo{person}{Saptarshi Ghosh}.}
  \bibinfo{year}{2017}\natexlab{}.
\newblock \showarticletitle{{Measuring Similarity among Legal Court Case
  Documents}}. In \bibinfo{booktitle}{\emph{Proceedings of the 10th Annual ACM
  India Compute Conference on ZZZ - Compute '17}}. \bibinfo{pages}{1--9}.
\newblock
\showISBNx{9781450353236}
\urldef\tempurl%
\url{https://doi.org/10.1145/3140107.3140119}
\showDOI{\tempurl}


\bibitem[\protect\citeauthoryear{Manning, Raghavan, and Schutze}{Manning
  et~al\mbox{.}}{2008}]%
        {Manning2008}
\bibfield{author}{\bibinfo{person}{Christopher~D. Manning},
  \bibinfo{person}{Prabhakar Raghavan}, {and} \bibinfo{person}{Hinrich
  Schutze}.} \bibinfo{year}{2008}\natexlab{}.
\newblock \bibinfo{booktitle}{\emph{{Introduction to Information Retrieval}}}.
  Vol.~\bibinfo{volume}{16}.
\newblock \bibinfo{publisher}{Cambridge University Press},
  \bibinfo{address}{Cambridge}. 100--103 pages.
\newblock
\showISBNx{9780511809071}
\showISSN{1351-3249}
\urldef\tempurl%
\url{https://doi.org/10.1017/CBO9780511809071}
\showDOI{\tempurl}


\bibitem[\protect\citeauthoryear{Mellinkoff}{Mellinkoff}{1963}]%
        {Mellinkoff1963}
\bibfield{author}{\bibinfo{person}{David Mellinkoff}.}
  \bibinfo{year}{1963}\natexlab{}.
\newblock \showarticletitle{{The language of the law}}.
\newblock \bibinfo{journal}{\emph{Boston: Little Brown and Company}}
  (\bibinfo{year}{1963}).
\newblock


\bibitem[\protect\citeauthoryear{Mikolov, Chen, Corrado, and Dean}{Mikolov
  et~al\mbox{.}}{2013}]%
        {Mikolov2013}
\bibfield{author}{\bibinfo{person}{Tomas Mikolov}, \bibinfo{person}{Kai Chen},
  \bibinfo{person}{Greg Corrado}, {and} \bibinfo{person}{Jeffrey Dean}.}
  \bibinfo{year}{2013}\natexlab{}.
\newblock \showarticletitle{{Efficient Estimation of Word Representations in
  Vector Space}}.
\newblock  (\bibinfo{year}{2013}), \bibinfo{pages}{1--12}.
\newblock
\showeprint{1301.3781}


\bibitem[\protect\citeauthoryear{Minocha, Singh, and Srivastava}{Minocha
  et~al\mbox{.}}{2015}]%
        {Minocha2015}
\bibfield{author}{\bibinfo{person}{Akshay Minocha}, \bibinfo{person}{Navjyoti
  Singh}, {and} \bibinfo{person}{Arjit Srivastava}.}
  \bibinfo{year}{2015}\natexlab{}.
\newblock \showarticletitle{{Finding Relevant Indian Judgments using Dispersion
  of Citation Network}}. In \bibinfo{booktitle}{\emph{Proceedings of the 24th
  International Conference on World Wide Web - WWW '15 Companion}}.
  \bibinfo{publisher}{ACM Press}, \bibinfo{address}{New York, New York, USA},
  \bibinfo{pages}{1085--1088}.
\newblock
\showISBNx{9781450334730}


\bibitem[\protect\citeauthoryear{Nanda, Siragusa, {Di Caro}, Boella, Grossio,
  Gerbaudo, and Costamagna}{Nanda et~al\mbox{.}}{2019}]%
        {Nanda2019}
\bibfield{author}{\bibinfo{person}{Rohan Nanda}, \bibinfo{person}{Giovanni
  Siragusa}, \bibinfo{person}{Luigi {Di Caro}}, \bibinfo{person}{Guido Boella},
  \bibinfo{person}{Lorenzo Grossio}, \bibinfo{person}{Marco Gerbaudo}, {and}
  \bibinfo{person}{Francesco Costamagna}.} \bibinfo{year}{2019}\natexlab{}.
\newblock \showarticletitle{{Unsupervised and supervised text similarity
  systems for automated identification of national implementing measures of
  European directives}}.
\newblock \bibinfo{journal}{\emph{Artificial Intelligence and Law}}
  \bibinfo{volume}{27}, \bibinfo{number}{2} (\bibinfo{year}{2019}),
  \bibinfo{pages}{199--225}.
\newblock
\showISBNx{0123456789}
\showISSN{15728382}
\urldef\tempurl%
\url{https://doi.org/10.1007/s10506-018-9236-y}
\showDOI{\tempurl}


\bibitem[\protect\citeauthoryear{Nickel and Kiela}{Nickel and Kiela}{2017}]%
        {Nickel2017}
\bibfield{author}{\bibinfo{person}{Maximilian Nickel} {and}
  \bibinfo{person}{Douwe Kiela}.} \bibinfo{year}{2017}\natexlab{}.
\newblock \showarticletitle{{Poincar{\'{e}} embeddings for learning
  hierarchical representations}}.
\newblock \bibinfo{journal}{\emph{Advances in Neural Information Processing
  Systems}} \bibinfo{volume}{2017-Decem}, \bibinfo{number}{Nips}
  (\bibinfo{year}{2017}), \bibinfo{pages}{6339--6348}.
\newblock
\showISSN{10495258}
\showeprint[arxiv]{1705.08039}


\bibitem[\protect\citeauthoryear{Ostendorff, Blume, and Ostendorff}{Ostendorff
  et~al\mbox{.}}{2020}]%
        {Ostendorff2020b}
\bibfield{author}{\bibinfo{person}{Malte Ostendorff}, \bibinfo{person}{Till
  Blume}, {and} \bibinfo{person}{Saskia Ostendorff}.}
  \bibinfo{year}{2020}\natexlab{}.
\newblock \showarticletitle{{Towards an Open Platform for Legal Information}}.
  In \bibinfo{booktitle}{\emph{Proceedings of the ACM/IEEE Joint Conference on
  Digital Libraries in 2020}}. \bibinfo{publisher}{ACM}, \bibinfo{address}{New
  York, NY, USA}, \bibinfo{pages}{385--388}.
\newblock
\showISBNx{9781450375856}


\bibitem[\protect\citeauthoryear{Ostendorff, Bourgonje, Berger,
  Moreno-Schneider, Rehm, and Gipp}{Ostendorff et~al\mbox{.}}{2019}]%
        {Ostendorff2019}
\bibfield{author}{\bibinfo{person}{Malte Ostendorff}, \bibinfo{person}{Peter
  Bourgonje}, \bibinfo{person}{Maria Berger}, \bibinfo{person}{Julian
  Moreno-Schneider}, \bibinfo{person}{Georg Rehm}, {and} \bibinfo{person}{Bela
  Gipp}.} \bibinfo{year}{2019}\natexlab{}.
\newblock \showarticletitle{{Enriching BERT with Knowledge Graph Embeddings for
  Document Classification}}. In \bibinfo{booktitle}{\emph{Proceedings of the
  15th Conference on Natural Language Processing (KONVENS 2019)}}.
  \bibinfo{publisher}{GSCL}, \bibinfo{address}{Erlangen, Germany},
  \bibinfo{pages}{305--312}.
\newblock


\bibitem[\protect\citeauthoryear{Pedregosa, Varoquaux, Gramfort, Michel,
  Thirion, Grisel, Blondel, Prettenhofer, Weiss, Dubourg, Vanderplas, Passos,
  Cournapeau, Brucher, Perrot, and Duchesnay}{Pedregosa et~al\mbox{.}}{2011}]%
        {scikit-learn}
\bibfield{author}{\bibinfo{person}{F. Pedregosa}, \bibinfo{person}{G.
  Varoquaux}, \bibinfo{person}{A. Gramfort}, \bibinfo{person}{V. Michel},
  \bibinfo{person}{B. Thirion}, \bibinfo{person}{O. Grisel},
  \bibinfo{person}{M. Blondel}, \bibinfo{person}{P. Prettenhofer},
  \bibinfo{person}{R. Weiss}, \bibinfo{person}{V. Dubourg}, \bibinfo{person}{J.
  Vanderplas}, \bibinfo{person}{A. Passos}, \bibinfo{person}{D. Cournapeau},
  \bibinfo{person}{M. Brucher}, \bibinfo{person}{M. Perrot}, {and}
  \bibinfo{person}{E. Duchesnay}.} \bibinfo{year}{2011}\natexlab{}.
\newblock \showarticletitle{Scikit-learn: Machine Learning in {P}ython}.
\newblock \bibinfo{journal}{\emph{Journal of Machine Learning Research}}
  \bibinfo{volume}{12} (\bibinfo{year}{2011}), \bibinfo{pages}{2825--2830}.
\newblock


\bibitem[\protect\citeauthoryear{Pennington, Socher, and Manning}{Pennington
  et~al\mbox{.}}{2014}]%
        {Pennington2014}
\bibfield{author}{\bibinfo{person}{Jeffrey Pennington},
  \bibinfo{person}{Richard Socher}, {and} \bibinfo{person}{Christopher
  Manning}.} \bibinfo{year}{2014}\natexlab{}.
\newblock \showarticletitle{{Glove: Global Vectors for Word Representation}}.
  In \bibinfo{booktitle}{\emph{Proceedings of the 2014 Conference on Empirical
  Methods in Natural Language Processing (EMNLP)}}. \bibinfo{publisher}{ACL},
  \bibinfo{address}{Stroudsburg, PA, USA}, \bibinfo{pages}{1532--1543}.
\newblock
\showISSN{00293970}
\urldef\tempurl%
\url{https://doi.org/10.3115/v1/D14-1162}
\showDOI{\tempurl}


\bibitem[\protect\citeauthoryear{Perozzi, Al-Rfou, and Skiena}{Perozzi
  et~al\mbox{.}}{2014}]%
        {Perozzi2014}
\bibfield{author}{\bibinfo{person}{Bryan Perozzi}, \bibinfo{person}{Rami
  Al-Rfou}, {and} \bibinfo{person}{Steven Skiena}.}
  \bibinfo{year}{2014}\natexlab{}.
\newblock \showarticletitle{{DeepWalk: online learning of social
  representations}}. In \bibinfo{booktitle}{\emph{Proceedings of the 20th ACM
  SIGKDD international conference on Knowledge discovery and data mining - KDD
  '14}}. \bibinfo{publisher}{ACM Press}, \bibinfo{address}{New York, New York,
  USA}, \bibinfo{pages}{701--710}.
\newblock
\showISBNx{9781450329569}
\showISSN{9781450329569}


\bibitem[\protect\citeauthoryear{Perozzi, Kulkarni, Chen, and Skiena}{Perozzi
  et~al\mbox{.}}{2017}]%
        {Perozzi2017}
\bibfield{author}{\bibinfo{person}{Bryan Perozzi}, \bibinfo{person}{Vivek
  Kulkarni}, \bibinfo{person}{Haochen Chen}, {and} \bibinfo{person}{Steven
  Skiena}.} \bibinfo{year}{2017}\natexlab{}.
\newblock \showarticletitle{{Don't Walk, Skip!: Online Learning of Multi-scale
  Network Embeddings}}. In \bibinfo{booktitle}{\emph{Proceedings of the 2017
  IEEE/ACM International Conference on Advances in Social Networks Analysis and
  Mining 2017}}. \bibinfo{publisher}{ACM}, \bibinfo{address}{New York, NY,
  USA}, \bibinfo{pages}{258--265}.
\newblock
\showISBNx{9781450349932}


\bibitem[\protect\citeauthoryear{Rehm, Bourgonje, Hegele, Kintzel, Schneider,
  Ostendor, Zaczynska, Berger, Grill, Rauchle, Rauenbusch, Rutenburg, Schmidt,
  Wild, Homann, Fink, Schulz, Seva, Quantz, Bottger, Matthey, Fricke, Thomsen,
  Paschke, Qundus, Hoppe, Karam, Weichhardt, Fillies, Neudecker, Gerber,
  Labusch, Rezanezhad, Schaefer, Zellhofer, Siewert, Bunk, Schlichting,
  Pintscher, Aleynikova, and Heine}{Rehm et~al\mbox{.}}{2020}]%
        {Rehm2020}
\bibfield{author}{\bibinfo{person}{Georg Rehm}, \bibinfo{person}{Peter
  Bourgonje}, \bibinfo{person}{Stefanie Hegele}, \bibinfo{person}{Florian
  Kintzel}, \bibinfo{person}{Julian~Moreno Schneider}, \bibinfo{person}{Malte
  Ostendor}, \bibinfo{person}{Karolina Zaczynska}, \bibinfo{person}{Armin
  Berger}, \bibinfo{person}{Stefan Grill}, \bibinfo{person}{Soren Rauchle},
  \bibinfo{person}{Jens Rauenbusch}, \bibinfo{person}{Lisa Rutenburg},
  \bibinfo{person}{Andre Schmidt}, \bibinfo{person}{Mikka Wild},
  \bibinfo{person}{Henry Homann}, \bibinfo{person}{Julian Fink},
  \bibinfo{person}{Sarah Schulz}, \bibinfo{person}{Jurica Seva},
  \bibinfo{person}{Joachim Quantz}, \bibinfo{person}{Joachim Bottger},
  \bibinfo{person}{Josene Matthey}, \bibinfo{person}{Rolf Fricke},
  \bibinfo{person}{Jan Thomsen}, \bibinfo{person}{Adrian Paschke},
  \bibinfo{person}{Jamal~Al Qundus}, \bibinfo{person}{Thomas Hoppe},
  \bibinfo{person}{Naouel Karam}, \bibinfo{person}{Frauke Weichhardt},
  \bibinfo{person}{Christian Fillies}, \bibinfo{person}{Clemens Neudecker},
  \bibinfo{person}{Mike Gerber}, \bibinfo{person}{Kai Labusch},
  \bibinfo{person}{Vahid Rezanezhad}, \bibinfo{person}{Robin Schaefer},
  \bibinfo{person}{David Zellhofer}, \bibinfo{person}{Daniel Siewert},
  \bibinfo{person}{Patrick Bunk}, \bibinfo{person}{Julia~Katharina
  Schlichting}, \bibinfo{person}{Lydia Pintscher}, \bibinfo{person}{Elena
  Aleynikova}, {and} \bibinfo{person}{Franziska Heine}.}
  \bibinfo{year}{2020}\natexlab{}.
\newblock \showarticletitle{{QURATOR: Innovative technologies for content and
  data curation}}. In \bibinfo{booktitle}{\emph{Proceedings of the Conference
  on Digital Curation Technologies (Qurator 2020)}}.
\newblock
\showeprint[arxiv]{2004.12195}


\bibitem[\protect\citeauthoryear{Rehm, Moreno-Schneider, Gracia, Revenko,
  Mireles, Khvalchik, Kernerman, Lagzdins, Pinnis, Vasilevskis, Leitner, Milde,
  and WeiÃŸenhorn}{Rehm et~al\mbox{.}}{2019}]%
        {rehm2019c}
\bibfield{author}{\bibinfo{person}{Georg Rehm}, \bibinfo{person}{Julian
  Moreno-Schneider}, \bibinfo{person}{Jorge Gracia}, \bibinfo{person}{Artem
  Revenko}, \bibinfo{person}{Victor Mireles}, \bibinfo{person}{Maria
  Khvalchik}, \bibinfo{person}{Ilan Kernerman}, \bibinfo{person}{Andis
  Lagzdins}, \bibinfo{person}{Marcis Pinnis}, \bibinfo{person}{Artus
  Vasilevskis}, \bibinfo{person}{Elena Leitner}, \bibinfo{person}{Jan Milde},
  {and} \bibinfo{person}{Pia WeiÃŸenhorn}.} \bibinfo{year}{2019}\natexlab{}.
\newblock \showarticletitle{{Developing and Orchestrating a Portfolio of
  Natural Legal Language Processing and Document Curation Services}}. In
  \bibinfo{booktitle}{\emph{Proceedings of Workshop on Natural Legal Language
  Processing (NLLP 2019)}}, \bibfield{editor}{\bibinfo{person}{Nikolaos
  Aletras}, \bibinfo{person}{Elliott Ash}, \bibinfo{person}{Leslie Barrett},
  \bibinfo{person}{Daniel Chen}, \bibinfo{person}{Adam Meyers},
  \bibinfo{person}{Daniel Preotiuc-Pietro}, \bibinfo{person}{David Rosenberg},
  {and} \bibinfo{person}{Amanda Stent}} (Eds.). \bibinfo{address}{Minneapolis,
  USA}, \bibinfo{pages}{55--66}.
\newblock
\newblock
\shownote{Co-located with NAACL~2019. 7 June 2019.}


\bibitem[\protect\citeauthoryear{Reimers and Gurevych}{Reimers and
  Gurevych}{2019}]%
        {Reimers2019}
\bibfield{author}{\bibinfo{person}{Nils Reimers} {and} \bibinfo{person}{Iryna
  Gurevych}.} \bibinfo{year}{2019}\natexlab{}.
\newblock \showarticletitle{{Sentence-BERT: Sentence Embeddings using Siamese
  BERT-Networks}}. In \bibinfo{booktitle}{\emph{The 2019 Conference on
  Empirical Methods in Natural Language Processing (EMNLP 2019)}}.
\newblock
\showeprint[arxiv]{1908.10084}


\bibitem[\protect\citeauthoryear{Rozemberczki, Kiss, and Sarkar}{Rozemberczki
  et~al\mbox{.}}{2020}]%
        {Rozemberczki2020}
\bibfield{author}{\bibinfo{person}{Benedek Rozemberczki},
  \bibinfo{person}{Oliver Kiss}, {and} \bibinfo{person}{Rik Sarkar}.}
  \bibinfo{year}{2020}\natexlab{}.
\newblock \showarticletitle{{An API Oriented Open-source Python Framework for
  Unsupervised Learning on Graphs}}.
\newblock  (\bibinfo{year}{2020}).
\newblock
\showeprint[arxiv]{2003.04819}


\bibitem[\protect\citeauthoryear{Salton, Wong, and Yang}{Salton
  et~al\mbox{.}}{1975}]%
        {Salton1975}
\bibfield{author}{\bibinfo{person}{G. Salton}, \bibinfo{person}{A. Wong}, {and}
  \bibinfo{person}{C.~S. Yang}.} \bibinfo{year}{1975}\natexlab{}.
\newblock \showarticletitle{{Vector Space Model for Automatic Indexing.
  Information Retrieval and Language Processing}}.
\newblock \bibinfo{journal}{\emph{Commun. ACM}} \bibinfo{volume}{18},
  \bibinfo{number}{11} (\bibinfo{year}{1975}), \bibinfo{pages}{613--620}.
\newblock
\showISSN{0001-0782}


\bibitem[\protect\citeauthoryear{Schwarzer, Schubotz, Meuschke, and
  Breitinger}{Schwarzer et~al\mbox{.}}{2016}]%
        {Schwarzer2016}
\bibfield{author}{\bibinfo{person}{Malte Schwarzer}, \bibinfo{person}{Moritz
  Schubotz}, \bibinfo{person}{Norman Meuschke}, {and} \bibinfo{person}{Corinna
  Breitinger}.} \bibinfo{year}{2016}\natexlab{}.
\newblock \showarticletitle{{Evaluating Link-based Recommendations for
  Wikipedia}}.
\newblock \bibinfo{journal}{\emph{Proc. of the 16th ACM/IEEE Joint Conference
  on Digital Libraries (JCDL`16)}} (\bibinfo{year}{2016}),
  \bibinfo{pages}{191--200}.
\newblock
\showISBNx{9781450342292}
\showISSN{15525996}


\bibitem[\protect\citeauthoryear{van Opijnen and Santos}{van Opijnen and
  Santos}{2017}]%
        {VanOpijnen2017}
\bibfield{author}{\bibinfo{person}{Marc van Opijnen} {and}
  \bibinfo{person}{Cristiana Santos}.} \bibinfo{year}{2017}\natexlab{}.
\newblock \showarticletitle{{On the concept of relevance in legal information
  retrieval}}.
\newblock \bibinfo{journal}{\emph{Artificial Intelligence and Law}}
  \bibinfo{volume}{25}, \bibinfo{number}{1} (\bibinfo{year}{2017}),
  \bibinfo{pages}{65--87}.
\newblock
\showISSN{15728382}


\bibitem[\protect\citeauthoryear{Vaswani, Shazeer, Parmar, Uszkoreit, Jones,
  Gomez, Kaiser, and Polosukhin}{Vaswani et~al\mbox{.}}{2017}]%
        {Vaswani2017}
\bibfield{author}{\bibinfo{person}{A. Vaswani}, \bibinfo{person}{N. Shazeer},
  \bibinfo{person}{N. Parmar}, \bibinfo{person}{J. Uszkoreit},
  \bibinfo{person}{L. Jones}, \bibinfo{person}{A.~N. Gomez},
  \bibinfo{person}{L. Kaiser}, {and} \bibinfo{person}{I. Polosukhin}.}
  \bibinfo{year}{2017}\natexlab{}.
\newblock \showarticletitle{{Attention Is All You Need}}.
\newblock \bibinfo{journal}{\emph{Advances in Neural Information Processing
  Systems}}  \bibinfo{volume}{30} (\bibinfo{date}{Jun} \bibinfo{year}{2017}),
  \bibinfo{pages}{5998--6008}.
\newblock


\bibitem[\protect\citeauthoryear{Wagh and Anand}{Wagh and Anand}{2020}]%
        {Wagh2020}
\bibfield{author}{\bibinfo{person}{Rupali~S. Wagh} {and} \bibinfo{person}{Deepa
  Anand}.} \bibinfo{year}{2020}\natexlab{}.
\newblock \showarticletitle{{Legal document similarity: A multicriteria
  decision-making perspective}}.
\newblock \bibinfo{journal}{\emph{PeerJ Computer Science}}
  \bibinfo{volume}{2020}, \bibinfo{number}{3} (\bibinfo{year}{2020}),
  \bibinfo{pages}{1--20}.
\newblock
\showISSN{23765992}
\urldef\tempurl%
\url{https://doi.org/10.7717/peerj-cs.262}
\showDOI{\tempurl}


\bibitem[\protect\citeauthoryear{Wang, Tan, and Han}{Wang
  et~al\mbox{.}}{2016}]%
        {Wang2016}
\bibfield{author}{\bibinfo{person}{Lidan Wang}, \bibinfo{person}{Ming Tan},
  {and} \bibinfo{person}{Jiawei Han}.} \bibinfo{year}{2016}\natexlab{}.
\newblock \showarticletitle{{FastHybrid: A hybrid model for efficient answer
  selection}}.
\newblock \bibinfo{journal}{\emph{Proceedings of the 26th International
  Conference on Computational Linguistics}} (\bibinfo{year}{2016}),
  \bibinfo{pages}{2378--2388}.
\newblock
\showISBNx{9784879747020}


\bibitem[\protect\citeauthoryear{Wiggers and Verberne}{Wiggers and
  Verberne}{2019}]%
        {Wiggers2019}
\bibfield{author}{\bibinfo{person}{Gineke Wiggers} {and} \bibinfo{person}{Suzan
  Verberne}.} \bibinfo{year}{2019}\natexlab{}.
\newblock \showarticletitle{{Citation Metrics for Legal Information Retrieval
  Systems}}. In \bibinfo{booktitle}{\emph{BIR@ECIR}}. \bibinfo{pages}{39--50}.
\newblock


\bibitem[\protect\citeauthoryear{Wikisource}{Wikisource}{2020}]%
        {WikiSource2020}
\bibfield{author}{\bibinfo{person}{Wikisource}.}
  \bibinfo{year}{2020}\natexlab{}.
\newblock \bibinfo{title}{{United States Supreme Court decisions by topic}}.
\newblock
\newblock
\urldef\tempurl%
\url{https://en.wikisource.org/wiki/Category:United_States_Supreme_Court_decisions_by_topic}
\showURL{%
\tempurl}


\bibitem[\protect\citeauthoryear{Williams, Nangia, and Bowman}{Williams
  et~al\mbox{.}}{2018}]%
        {Williams2018}
\bibfield{author}{\bibinfo{person}{Adina Williams}, \bibinfo{person}{Nikita
  Nangia}, {and} \bibinfo{person}{Samuel Bowman}.}
  \bibinfo{year}{2018}\natexlab{}.
\newblock \showarticletitle{{A Broad-Coverage Challenge Corpus for Sentence
  Understanding through Inference}}.
\newblock  (\bibinfo{year}{2018}), \bibinfo{pages}{1112--1122}.
\newblock
\urldef\tempurl%
\url{https://doi.org/10.18653/v1/n18-1101}
\showDOI{\tempurl}


\bibitem[\protect\citeauthoryear{Winkels, Boer, Vredebregt, and {Van
  Someren}}{Winkels et~al\mbox{.}}{2014}]%
        {Winkels2014}
\bibfield{author}{\bibinfo{person}{Radboud Winkels}, \bibinfo{person}{Alexander
  Boer}, \bibinfo{person}{Bart Vredebregt}, {and} \bibinfo{person}{Alexander
  {Van Someren}}.} \bibinfo{year}{2014}\natexlab{}.
\newblock \showarticletitle{{Towards a Legal Recommender System}}. In
  \bibinfo{booktitle}{\emph{Frontiers in Artificial Intelligence and
  Applications}}, Vol.~\bibinfo{volume}{271}. \bibinfo{pages}{169--178}.
\newblock
\showISBNx{9781614994671}
\showISSN{09226389}


\end{thebibliography}

\extended{}{\vfill

\pagebreak

\appendix

\section{Detailed Case Descriptions and Comparisons}
\label{asec:manual}

\paragraph{Seed decision: Mugler v Kansas.}

A new Kansas law prohibited the sale and manufacture of intoxicating liquor. Prior to the passage of the Kansas law, Mugler built a brewery. Mugler was indicted for violating the law and having manufactured intoxicating liquors without a permit. The main issue is if the Kansas law violated the Due Process Clause of the Fourteenth Amendment. More specifically, does prohibiting the sale and manufacture of intoxicating liquors, subsequently lowering the economic value of property, deprive the owner of that property and as articulated in the Due Process Clause of the Fourteenth Amendment? 

The court decided that the Kansas law does not infringe on Fourteenth Amendment rights or privileges. It stated that the principle requiring property holders not to use their property so as to be injurious to the community was compatible with the Fourteenth Amendment. Moreover, the court reasoned that a prohibition on the use of property, by valid legislation, for purposes of protecting the health and safety of the community, cannot be deemed a taking or an appropriation of property for public benefit. Since the legislation did not restrict the owner's control, right to dispose, or ability to use property for lawful purposes, no taking had occurred. If the legislature needs to act due to public safety, it cannot discontinue such activity because individuals suffer inconveniences.

\paragraph{Yick Wo v Hopkins.} %

A San Francisco ordinance required all laundries in wooden buildings to hold a permit issued by the city's Board of Supervisors. The board had total discretion over who would be issued a permit. The majority of laundry businesses were operated by Chinese workers, but not a single Chinese owner was granted a permit. Yick Wo and Wo Lee, who  operated a laundry business without a permit, were imprisoned after refusing to pay a fine. They sued for habeas corpus and argued that discriminatory enforcement of the ordinance violated their rights under the Equal Protection Clause of the Fourteenth Amendment. 

The Supreme Court of California and the Circuit Court of the United States for the District of California denied the claims. The main problem was if the unequal enforcement of the ordinance violates Yick Wo and Wo Lee's rights under the Equal Protection Clause of the Fourteenth Amendment? The Court concluded that, despite the impartial wording of the law, its biased enforcement violated the Equal Protection Clause and therefore violated the provision of the Fourteenth Amendment. The judgment of the Supreme Court of California and Circuit Court of the United States for the District of California were reversed, and the cases remanded. 

Like the seed decision, the main problem of this case is a state law that allegedly infringes the Fourteenth Amendment. While the seed decision focuses more on the Due Process Clause, this case addresses the Equal Protection and Citizenship Clause. So we say it is \textit{not related}.

\paragraph{Munn v Illinois.}  %

The legislature of Illinois regulated grain warehouses and elevators by establishing maximum rates that private companies could charge for their use and storage of agricultural products. The grain warehouse firm Munn and Scott was found guilty of violating the law. The company appealed the conviction on the grounds that the law was an unconstitutional deprivation of property without due process of law and that the rates deny the warehouse equal protection that violated the Fourteenth Amendment. The court ruled in favor of the State. It argued that the states can regulate the use of private property when the regulation is necessary for the public good. Moreover, the court declared that even though interstate commerce is the responsibility of Congress, a state could take action in the public interest without impairing the federal control.

Similar to the seed decision, the main problem of this case is a state law that allegedly infringes the Fourteenth Amendment. Like the seed decision, this case addresses the violation of the Due Process Clause of the Fourteenth Amendment. In both cases the court argued that individual interests outweigh public interests which justifies the regulations. This led both cases to be ruled in favor of the state. The case is \textit{related}.

\paragraph{Lifestock Dealers Butchers v Crescent City Lifestock 1870.} %

An act passed by the legislature of the State of Louisiana prohibited all persons and corporations to land, keep, or slaughter any animals at any place within the city and parishes of New Orleans. Only the company created and organized under the new act, the "Crescent City Live-stock Landing and Slaughter Company" was entitled to do the aforementioned. The act was passed on March 1869 and was described as an act to protect the health of the city of New Orleans. A group of excluded butchers sought an injunction against the monopoly on the grounds that they were prevented from practising their trade. 

The state courts upheld the law. The appeal was based on the following grounds: the act created an involuntary servitude forbidden by the Thirteenth Amendment, it abridges the privileges and immunities of citizens of the U.S., it denied plaintiffs the equal protection of the laws and deprived them of their property without due process of law, which is all protected under the Fourteenth Amendment. The court stated the involuntary servitude of the Thirteenth Amendment is restricted to personal servitude, not a servitute attached to property and that only privileges and immunities of U.S. citizens are protected by the Fourteenth Amendment so that those of state citizens are unaffected. Moreover, the equal protection clause of the Fourteenth Amendment is primarily intended to prevent discrimination against blacks. The court concluded that the prohibition of the plaintiffs' trade cannot be held to be a deprivation of property with regard to the Fourteenth Amendment. This case was the first case requiring interpretation of the amendments. 

Similar to the seed decision, the Court had to interpret and apply due process for a regulation in the public interest. The case is \textit{related}.

\paragraph{Butchers Benevolent Crescent City Lifestock 1872.}  %

This is another opinion with the same background as the previous case (see Lifestock Dealers Butchers v Crescent City Lifestock 1870), but with a different plaintiff. Again the decision of the court rules that the Fourteenth Amendment did not forbid Louisiana's use of its police powers to regulate butchers. The Court held that the Fourteenth Amendment's Privileges or Immunities Clause affected only rights of U.S. citizenship. Therefore, according to the court, the butcher's Fourteenth Amendment rights had not been violated. 

As before, this is \textit{related}.

\paragraph{Lochner v New York.} %

The state of New York enacted the Bakeshop Act, a statute which forbade bakers to work more than 60 hours a week / 10 hours a day. Lochner was accused of permitting an employee to work more than 60 hours in one week. He was charged with fines. Lochner appealed but lost in state court. He argued that the Fourteenth Amendment should have been interpreted to contain the freedom of contract among the rights emcompassed by substantive due process. In his view the right to purchase or to sell labor should be part of the liberty protected by the amendment. 

The question that arises is if the Bakeshop Act violates the liberty protected by the Due Process Clause of the Fourteenth Amendment. The court invalidated the New York statute on the grounds, that it interfered with the freedom of contract and therefore the Fourteenth Amendment's right to liberty afforded employer and employee. Moreover the New York statute failed the rational basis test for determining whether the government action is constitutional. The majority reasoned that the Bakeshop Act had no rational basis because long working hours did not dramatically undermine employees' health and baking is not dangerous per se.  

Same as in the seed decision, the court said that the power of the courts to review legislative action in a matter affecting the general welfare exists only when a statute enacted to protect the public health or safety has no real or substantial relation to those objects, or is a plain invasion of rights secured by the Fourteenth Amendment. Different from the seed decision, however, the court found the enacted New York statute to have no rational basis and ruled in favor of the plaintiff. This case is \textit{related}.

\paragraph{Allgeyer v. Louisiana.} %

A Louisiana statute prohibited out-of-state insurance companies from conducting business in Louisiana whithout maintaining at least one place of business and authorized agent within state. The intention behind the implementation of the statute was that it protects citizens from deceitful insurance companies. Allgeyer \& Company violated the statute by purchasing insurance from a New-York-based company. The issue was whether the Louisiana statute violates the Fourteenth Amendment's Due Process Clause, which protects companies' liberty to enter in to contracts with businesses of their own choice. 

The court ruled in favor of the plaintiff and found that the Louisiana statute deprived Allgeyer \& Company of its liberty without Due Process under the Fourteenth Amendment. Moreover, it found that the Fourteenth Amendment extends to protect individuals from restrictions of their freedom to contract in pursuit of one's livelihood or vacation. 

Unlike in the seed decision, the Supreme Court of the United States chose to analyze the possible violation of the Fourteenth Amendment from the standpoint of the person rather than the company. The state maintains policing power in relationship to the company, but it cannot legislate in such manner as to deny an individual's liberty. In the seed decision however, the court decided that public health and safety is to prioritize over the individual. This is \textit{related}.

\paragraph{Calder v. Wife.} %

A Connecticut probate court denied Mr. and Mrs. Caleb Bull (the stated beneficiaries of Norman Morrison's will) an inheritance. When the Bulls wanted to appeal the decision more than 1,5 years later, they found that a state law prohibited appeals not made within 18 months of the ruling. The Bulls persuaded the Connecticut legislature to change the restriction, which enabled them to successfully appeal the case. Calder, the initial inheritor of Morrison's estate, took the case to the Supreme Court. The main issue was, if the Connecticut legislation violates Art. 1 Section 10 of the Constitution, which prohibits ex post facto laws. 

The court decided that the Connecticut legislation was not an ex post facto law arguing that restrictions against ex post facto laws were not designed to protect citiziens' contract rights but only criminal matters. Moreover, all ex post facto laws are retrospective, but all retrospective laws are not necessarily ex post facto and even vested property rights are subject to retroactive laws.

This case is \textit{not related}.

\paragraph{Davidson v. New Orleans.} %

The city of New Orleans sought to make an assessment on certain real estate within the the parishes of Carroll and Orleans for the purpose of draining swamp lands there. A part of John Davidson's estate was included in the assessment and was assessed for \$50,000. The main issue was whether Mrs. Davidson (widow of Mr. Davidson) was being deprived of her property without due process of law clause of the Fourteenth Amendment. 

The court ruled against the plaintiff. The court stated that whenever a state takes property for public use, and state laws provide a mode for contesting the charge in the ordinary courts, and if due notice is given to the person, and if there is a full and fair hearing, there is no cause for a suit charging lack of due process of the law. Moreover, the court said that a due process of law does not imply a regular proceeding in a court of justice and that the Fourteenth Amendment was not being infringed.

Similar to the seed case, this case discusses the Due Process Clause of the Fourteenth Amendment. Like in the seed case the state argued that whenever by the laws of a state, or by state authority a burden is imposed upon property for the public use, with notice to the person and/or adequate compensation, it cannot be said to deprive the owner of this property without due process of law. This is \textit{related}.

\paragraph{Muller v. Oregon.} %

Oregon enacted a law that limited women to 10 hours of work in factories and laundries. Curt Muller, the owner of a laundry business, was fined when he violated the law. Muller appealed the conviction. The main issue was whether the Oregon law violated the Fourteenth Amendment. 

The court upheld Oregon law. Even though the case Lochner v. New York dealt with the same issues of limiting work hours, the court distinguished this case because of the  existing difference between the sexes. Furthermore, the court reasoned that the child-bearing nature and social role of women provided a strong state interest in reducing their working hours. 

Similar to in the seed case, the court found the enacted Oregon law to have rational basis since the law protects public health and therefore does not violate the Fourteenth Amendment. \textit{Related}.

\paragraph{Kidd v Pearson.} %

An Iowa state law made the manufacture of liquor in the state illegal, even when the liquor was for sale and consumption out-of-state. The main issue was whether or not the state law was in conflict with the power of Congress to regulate interstate commerce. 

The Court decided that there is no conflict and the state law is valid. The Court erected a distinction between manufacture and commerce. The state law regulated manufacturing only. The justices feared that a broad view of commerce that would embrace manufacturing would also embrace the power to regulate every step of industry. The court ruled that there was not a conflict between Congress' power to regulate interstate commerce and the state law covering manufacturing within a given state. Therefore, the law was valid.

This case is different from the seed case in its focus on interstate commerce rather than due process. But it discusses similar issues as in the seed case. In the seed case the court decided that a state has the right to prohibit or restrict the manufacture of intoxicating liquors within her limits; to prohibit all sale and traffic in them in said State; to inflict penalties for such manufacture and sale, and to provide regulations for the abatement as a common nuisance of the property used for such forbidden purposes; and that such legislation by a State is a clear exercise of her undisputed police power, which does not abridge the liberties or immunities of citizens of the United States, nor deprive any person of property without due process of law, nor in any way contravenes any provision of the Fourteenth Amendment to the Constitution of the United States. In this case the court agreed with the decision of the lead case and ruled similarly in that matter. It is \textit{related}, mostly on factual grounds but also it is somewhat legally related.

\paragraph{Lawton v Steele.} %

A New York statute preserved fisheries from extractive and exhaustive fishing. It said that nets set upon waters of the state or on the shores of or islands in such waters in violation of the statutes of the state enacted for the protection of fish, may be summarily destroyed by any person and asked certain officers to remove them. Steele, a game and fish protector, removed nets of the alleged value of \$525 belonging to the plaintiff. 

The taking and destruction of the nets were claimed to have been justifiable under the statutes of the state relating to the protection of game and fish. Plaintiffs claimed there was no justification under the statutes, and if they constituted such justification upon their face, they were unconstitutional. The court decided in favor of the defendant, and held the New York statute to be constitutional. 

Similar to the seed case, this case discusses whether or not the Fourteenth Amendment was violated with regard to the Due Process Clause. This case is \textit{related}.

\paragraph{Geer v Connecticut.} %

A Connecticut statute provided that it is prohibited to kill woodcook, ruffled grouse, and quail for conveyance across  state borders. Geer was convicted of possessing woodcock, ruffled grouse, and quail with the unlawful intent of transporting them out of state. 

The Court concluded that the state had the right to keep the game birds within the state for all purposes and to create and regulate its own internal state commerce with respect to the birds. Therefore the statute did not violate the Constitution. The Court explained that the state had the police power to preserve a food supply that belonged to the people of Connecticut by requiring that the commerce in game birds be kept within the state.

Similar to the seed case, the court decided a due process case with regard to the benefit of the people of the state. \textit{Related}.

\paragraph{Groves v Slaughter.}  %

A provision of the Mississippi constitution disallowed bringing slaves into the state for sale. Slaughter took a group of slaves to Mississippi to sell them. He accepted partial payment. The note fell due but remained unpaid. A federal court eventually held that Slaughter was entitled to recover the amount of the contract. This prohibition was challenged as being an unlawful restriction of interstate commerce in violation of the Commerce Clause. The provision did not become effective until a supporting statute was enacted, but that supporting statute followed the sale in question. Hence, the court decided that the contract was valid.

This is somewhat related to the small part of the seed case related to the Interstate Commerce Clause. But it is unrelated to the main focus on Due Process. \textit{Unrelated}.

\paragraph{Rast v. Van Deman.} %

A Florida statute of 1913 imposed special license taxes on merchants using profit sharing coupons and trading stamps. A suit was brought to restrain the enforcement of the statue on the ground that it violates the contract and the commerce clauses and the due process and equal protection provisions of the Fourteenth Amendment. 

The court decided that the statute does not offend any constitutional provisions but held that the statute showed that the conditions of complainant's business and property engaged therein are such that enforcement of the statute would produce irreparable injury, it furnishes ground for equitable relief.

Like the seed case, this case addresses the violation of the Due Process Clause and Equal Protection Clause of the Fourteenth Amendment. This is about taxes, rather than regulation, though, so it is \textit{unrelated}.

\paragraph{County of Mobile v. Kimball.} %

An act created a board of commissioners for the improvement of the river, harbor, and bay of Mobile, and required the president of the commissioners of revenue of Mobile County to issue bonds to the amount of \$1,000,000, and deliver them, when called for, to the board, to meet the expenses of the work directed. The board was authorized to apply the bonds, or their proceeds, to the cleaning out, deepening, and widening of the river, harbor, and bay of Mobile, or to the construction of an artificial harbor in addition to such improvement. The board of commissioners entered into a contract with the complainants, Kimball and Slaughter, to dredge and cut a channel through a designated bar in the bay. The work agreed upon was completed and accepted by the board through its authorized engineer. The amount due to them was not fully paid. 

The court decided that the act of the Legislature of Alabama is invalid, as it conflicts with the commercial power vested in Congress. 

This case is on an \textit{unrelated} issue.

\paragraph{Brass v. ND. Ex Rel. Stoeser.} %

A North Dakota state law defined persons operating grain elevators as public warehouse men and regulated their fees and charges. Brass, such an operator, refused to receive certain grain at the storage charges provided by the law, alleging they were too low, and a writ of mandate was issued out of the State Court to require him to do so. 

The court affirmed, holding that the power of the State to regulate the grain elevator business did not depend upon the fact of a practical monopoly by the elevator owners. The Court held the law to be constitutional under which the elevator operator was required to make contracts at fees and charges under conditions. 

The main issue is upon whether or not the Congress may legislate commercial power. The court decided that the harbor board, created by a law of the State, was authorized to make contracts for a public work in which the county was specially interested, and by which it would be immediately and directly benefited, and to require obligations of the county to meet the expenses incurred. Furthermore, the court argued that it is enough that by force of the law of its creation it could bind the county for work for which it contracted. Having thus bound the county, the contractors are entitled to the bonds stipulated, or their equivalent in money.

Like the seed case, this case addresses the violation of the Due Process Clause and Equal Protection Clause of the Fourteenth Amendment. It is \textit{related}.

\paragraph{Erie R. Co. v. Williams.} %

The contention of plaintiff is that the Labor Law is repugnant to the Fourteenth Amendment because it deprives the company of property and the employees of liberty without due process of law. The court decided that the law operates not only to require the railroads to pay their employees semi-monthly, but prohibits them from making contracts with their employees which shall vary the time of payment. 

The court rejected both contentions of plaintiff and sustained the law as an exercise of the power over plaintiff's charter; and  that the requirement of semi-monthly payments was an unconstitutional interference with interstate commerce. The Supreme Court affirmed the previous decision. 

Similar to the seed case, this case discusses whether or not the Fourteenth Amendment was violated with regard to the Equal Protection Clause and Due Process Clause. This case is \textit{related}.

\paragraph{Hall v. Geiger-Jones Co.} %

The Ohio blue sky law is a restraint upon the disposition of certain property, and requires dealers in securities evidencing title to or interest in such property to obtain a license. Under the blue sky laws, brokers who sold  securities within Ohio were to be licensed to do so. To obtain a license, a designated executive officer needed to be satisfied of the good business repute of the applicants and their agents, and licenses, when issued, could be revoked by him upon ascertaining that the licensees were of bad business repute, violated any provision of the act, or engaged in illegitimate business or fraudulent transactions. Appellee Geiger-Jones Co. filed an action seeking to enjoin enforcement of Ohio's blue sky laws. The district granted the injunctive relief. Hall appealed. 

The main question was whether or nor the Ohio's blue sky laws were properly enjoined. The Supreme Court reversed the district court's judgment and remanded the matter for further proceedings. The Court ruled that the powers conferred to Hall were not arbitrary or violative of the due process clause of Fourteenth Amendment. Moreover, the blue sky laws did not interfere with interstate commerce and, therefore, did not violate the commerce clause. According to the Court, such regulation affected interstate commerce in securities only incidentally.

Similar to the seed case, this case discusses whether or not the Fourteenth Amendment was violated with regard to the Due Process Clause. It is \textit{related}.

}

\end{document}